\documentclass[accepted]{uai2026} 
                        
\usepackage[american]{babel}

\usepackage{natbib} 
\usepackage{xr}
\usepackage{xr-hyper}
\usepackage{floatflt}

\usepackage{amsmath}
\usepackage{amssymb}
\usepackage{bbm}
\usepackage{amsthm}
\usepackage{mathtools}
\usepackage{subcaption, graphicx}
\usepackage{xcolor}
\usepackage{wrapfig}
\usepackage{pifont} 
\usepackage{nicefrac}       
\usepackage{amsfonts}       
\usepackage{booktabs}       
\usepackage{url}            
\usepackage[T1]{fontenc}    
\usepackage[utf8]{inputenc} 
\usepackage{glossaries}
\usepackage{float}
\usepackage{algorithm}    
\usepackage{algorithmic}  
\usepackage[framemethod=tikz]{mdframed}
\usepackage[most]{tcolorbox}
\tcbuselibrary{skins,breakable}
\usepackage{caption}
\usepackage{chemarrow}
\usepackage{siunitx}
\usepackage{multirow}
\usepackage{hyperref}

\makeatletter
\def\@icmlfirsttime{0}  
\makeatother

\tcbuselibrary{breakable, skins} 

\usepackage{scalerel}
\usepackage{mleftright}
\usepackage{bm}

\DeclareMathOperator{\E}{{}\mathbb{E}}

\renewcommand{\to}{\ensuremath{\rightarrow}}              

\newcommand{\grpP}[1]{\ensuremath{\mleft( #1 \mright)}}   
\newcommand{\fnP}[2]{\ensuremath{#1 \grpP{#2}}}           

\newcommand{\p}[1]{\fnP{p_{{\theta}}}{#1}}

\newcommand{\mi}{I}

\newcommand{\expect}{\mathbb{E}}	
\newcommand{\R}{\mathbb{R}}

\usepackage[capitalize,noabbrev]{cleveref}
\crefformat{section}{§#2#1#3}
\crefformat{subsection}{§#2#1#3}

\theoremstyle{plain}
\newtheorem{theorem}{Theorem}[section]
\newtheorem{proposition}[theorem]{Proposition}

\theoremstyle{definition}

\theoremstyle{remark}

\definecolor{mygray}{RGB}{240,240,240}

\usepackage{tikz} 

\title{Optimizing Likelihoods via Mutual Information: Bridging Simulation-Based Inference and Bayesian Optimal Experimental Design}

\author[1]{\href{mailto:<vzaballa@uci.edu>?Subject=Your UAI 2026 paper}{Vincent D. Zaballa}{}}
\author[1]{Elliot E. Hui}
\affil[1]{%
    Department of Biomedical Engineering\\
    University of California, Irvine\\
    Irvine, California, USA
}
  
\begin{document}
\maketitle

\begin{abstract}
  Simulation-based inference (SBI) typically depends on expensive simulators, so inference and experimental design must operate under fixed simulation budgets. Bayesian optimal experimental design (BOED) maximizes expected information gain (EIG), but this utility function can hide shortcomings of the underlying inference objects that will be used in downstream inference tasks (posterior estimation).  We show that the InfoNCE lower bound on EIG becomes a principled SBI training objective for conditional normalizing flow density model - the workhorse of SBI applications - with an asymptotic likelihood-fitting interpretation that links MI maximization to surrogate-likelihood learning and thereby increases the value of each simulator call for both inference and design. We propose SBI-BOED, a single stochastic-gradient procedure that jointly trains a conditional normalizing-flow likelihood and optimizes designs without requiring simulator differentiability or pathwise gradients to experimental designs. The same MI view also yields simulation active learning under fixed designs via expected predictive information gain (EPIG). With an InfoNCE-$\lambda$ objective and practical stabilizers, SBI-BOED improves posterior calibration and predictive accuracy over strong BOED baselines on synthetic benchmarks and scientific simulators at matched simulation budgets.
\end{abstract}

\section{Introduction}
\label{sec:introduction}

Scientific models can be described as simulators that simulate complex, potentially stochastic, interactions that yield observations from the interactions. Additionally, these simulators sometimes have \emph{controllable} inputs that yield a different distribution of observations. For example, a simulator generates an output $y$ based on experimental designs $\xi$ (controllable inputs) and model parameters $\theta$ (latent system properties that govern the input-output relationship). Inferring the posterior distribution of model parameters given observed data $p(\theta|y,\xi)$ represents a fundamental challenge in Bayesian statistics and can be viewed as solving an inverse problem for a given simulator \cite{lindley1972}. In Simulation-Based Inference (SBI), \emph{some} simulators implicitly define a design-dependent likelihood $p(y|\theta,\xi)$ that, combined with a prior over model parameters $p(\theta)$, enables inference of the posterior distribution $p(\theta|y_o,\xi)$ where $y_o$ represents observed data  \citet{cranmer2020frontier}. SBI methods address scenarios where the likelihood is intractable by approximating either 1) the likelihood $p(y|\theta,\xi)$ or 2) the posterior $p(\theta|y,\xi)$ directly using neural density estimators, or, 3) using classifiers to estimate the likelihood-to-evidence ratio, $\frac{p(y,\theta|\xi)}{p(\theta)p(y|\xi)}$. Importantly, simulations can be costly  (i.e. \emph{slow}) to collect, introducing a bottleneck in downstream applications of simulators.

Inferring the parameters of the model given observations requires performing experiments that may be costly to perform. For example, in drug development, collecting data is expensive and is partially responsible for the great cost associated with developing new therapies where thousands to millions of designs may be employed \emph{per round} of experimentation in a high-throughput screening campaign \citep{paul2010improve}. Triaging which data points to gather can help reduce the time to develop a new therapy for patients. It is thus imperative to prioritize collection of observed data $y_o$, using optimal designs $y_o|\xi^*$, to arrive at an accurate and \emph{calibrated} inference of model parameters to make predictions of future responses, such as the likelihood of successful drug treatment.

Towards reducing cost, Bayesian optimal experimental design (BOED) has shown promise as a method for optimizing experiments $\xi$, even when dealing with potentially high-dimensional design spaces. This can be achieved by employing a likelihood model along with priors for the parameters of interest \citep{lindley1956,Foster2019, kleinegesse2019efficient}. BOED operates by both \emph{evaluating} and \emph{optimizing} a utility function of the Expected Information Gain (EIG) upon performing a proposed experiment \citet{lindley1956, rainforth2024modern}.
\begin{equation}
    I(\xi) \triangleq \E_{p(y|\xi)}\left[ H[p(\theta)] - H[p(\theta|y,\xi)] \right].
\end{equation}

The EIG  can be used as an approximation for the information gained in an experiment with design $\xi$. The intuition is: which experimental design and outcome would be most surprising given what we assume about the model when conducting the experiment? This would be the optimal experimental design and can be rewritten into the form of calculating the mutual information (MI), where $\text{MI}(\theta;y|\xi) = \mi(\xi)$, between the observed data and unknown parameters as the ratio of likelihood to marginal likelihood or posterior to prior
\begin{align}
\begin{split}
    \text{MI}(\theta;y|\xi) & = \expect_{p(\theta)p(y|\theta,\xi)}\left[\log \frac{p(y|\theta,\xi)}{p(y|\xi)} \right] \\
    & = \expect_{p(\theta)p(y|\theta,\xi)}\left[\log \frac{p(\theta| y,\xi)}{p(\theta)} \right].
    \label{eq:infoGain}
\end{split}
\end{align}

However, most BOED utility functions require either an explicit likelihood or a differentiable one. Previous BOED work focused on estimating the MI as an objective function within an outer optimizer, such as Bayesian optimization, which results in a nested optimization process \citep{gutmann2016bayesOptLFI, rainforth2018nestingmontecarloestimators, Kleinegesse2020b, Foster2019}. Notably, these methods can be combined with implicit models such as those used in SBI, but they can be inefficient, motivating methods to simultaneously optimize the design and MI in a single optimization process \citep{Foster2019a}. However, unified optimization typically depends on an \textit{explicit} likelihood or an implicit likelihood with a \textit{differentiable} simulator \citep{Kleinegesse2021,ivanova2021implicit}, except for \emph{simulation-heavy} reinforcement learning (RL) methods \cite{lim2022policybased, blau2022optimizingsequentialexperimentaldesign, blau2024statisticallyefficientbayesiansequential}, which are often not feasible in SBI settings where generating new simulator samples takes seconds or longer.

Expensive simulators also motivate an active-learning perspective on SBI: under a fixed simulation budget, not all simulator queries are equally valuable. Rather than drawing simulations uniformly, one can prioritize parameter settings whose outcomes are expected to maximally improve the learned likelihood or posterior surrogate. In our approach, this prioritization follows from the same InfoNCE-regularized mutual-information objective used for design optimization, yielding a unified principle for both selecting informative experiments and allocating simulator calls. This further reduces simulation burden by focusing computation on the most informative queries.


We show that SBI and BOED can benefit from a shared information-theoretic perspective. On a standard SBI benchmark, we first study how InfoNCE and its regularized variant affect the training dynamics of amortized likelihood models. We then use the same bound to define the \emph{value} of a simulation, yielding an active-learning strategy for selecting new latent parameters $\theta$ under a fixed simulation budget. Next, we apply this objective to BOED with implicit models, enabling design optimization and surrogate training without requiring a differentiable simulator and avoiding the simulation cost of RL-based methods. Finally, we show that SBI-inspired iterative inference can improve cumulative information gain and posterior calibration over multiple rounds of BOED. Overall, our results highlight the value of cross-pollination between SBI and BOED for improving both inference quality and simulation efficiency. We summarize the offline simulator-call comparison in \cref{tab:simulator_calls}.

\begin{table}[t]
\vspace{2mm}
\caption{Offline simulator calls to learn an amortized design strategy; one call is one draw $y\sim p(y\mid\theta,\xi)$. On SIR (\cref{sec:SIR_model}), SBI-BOED exceeds iDAD accuracy while using $\sim$50$\times$ fewer simulator calls.}
\label{tab:simulator_calls}
\centering
\small
\setlength{\tabcolsep}{4pt}
\renewcommand{\arraystretch}{1.15}
\begin{tabular}{lccc}
\toprule
& RL-BOED$^\dagger$ & iDAD & \textcolor{black!20!blue}{SBI-BOED} \\
\midrule
Simulator calls & $\sim 10^{7}\text{--}10^{8}$ & $\approx 2.6\times 10^{8}$ & \textcolor{black!20!blue}{$\approx 5.1\times 10^{6}$} \\
\bottomrule
\end{tabular}

\vspace{0.8mm}
\footnotesize $^\dagger$Not evaluated on SIR; Blau et al.\ report training amortized RL-BOED agents with tens of millions of simulated experiments.
\vspace{-4mm}
\end{table}

In bridging BOED and SBI, our key contributions are:
\begin{itemize}
    \item \textbf{Regularized MI Bound for Improved Inference}: We show that the $I_{\text{NCE-}\lambda}$ utility improves calibration and predictive accuracy relative to the standard InfoNCE estimator, and we analyze its effect on training dynamics for amortized likelihood models.
    
    \item \textbf{Active Learning for Simulation-Efficient SBI}: We propose an active-learning strategy for selecting new latent parameters $\theta$ under a fixed simulation budget, using the InfoNCE-regularized mutual-information bound to prioritize simulator queries that are expected to be most informative.
    
    \item \textbf{Practical Optimization for Implicit Likelihoods}: We develop a robust method to jointly optimize experimental designs and train SBI surrogates without requiring a differentiable simulator. To address the limitations of naive gradient-based design optimization, we optimize a design \emph{distribution}, improving performance in non-differentiable settings.
    
    \item \textbf{Multi-Round BOED with SBI Principles}: We show that incorporating SBI ideas such as repeated rounds of inference improves cumulative information gain and yields better-calibrated posteriors over multiple rounds of BOED.
\end{itemize}

\section{Background}
\label{sec:background}

\subsection{Simulation-Based Inference}

SBI addresses Bayesian inference in settings where a simulator $g$ can generate synthetic observations but the likelihood is unavailable or impractical to evaluate. Given a prior $\theta \sim p(\theta)$ and (optionally) a design $\xi$, the simulator defines a data-generating process $y = g(\theta,\xi,z)$ with randomness $z$, inducing an implicit likelihood $p(y\mid\theta,\xi)$ and posterior $p(\theta\mid y_o,\xi)$. Modern SBI methods learn amortized surrogates from simulated pairs $(\theta,y)$ by training neural density estimators (e.g., normalizing flows) to approximate either the likelihood $p_\phi(y\mid\theta,\xi)$ or the posterior $p_\phi(\theta\mid y,\xi)$ via maximum likelihood \citep{Papamakarios2019, lueckmann2021benchmarking}. Alternatively, SBI can estimate the likelihood-to-evidence ratio by training a classifier to distinguish samples from the joint $p(\theta,y\mid\xi)$ versus the product of marginals $p(\theta)p(y\mid\xi)$, yielding $\exp f_\phi(\theta,y)\approx \frac{p(y\mid\theta,\xi)}{p(y\mid\xi)}$ (and related ratios). The choice between posterior, likelihood, or ratio-based SBI depends on downstream needs, e.g., amortized posteriors are convenient for repeated inference on many observations, while ratio ensembles can improve robustness (e.g., under calibration tests) at additional compute cost \citep{talts2020validating, hermanscrisis}. 

\textbf{Neural Likelihood Estimation} \hspace{0.5em} We can use data from the joint distribution to train a conditional neural density-based likelihood function (NL). If we take a dataset of samples $\{ y_n, \theta_n \}_{1:N}$ obtained from a simulator as previously described, we can train a conditional density estimator $p_\phi (y|\theta)$ to model the likelihood by maximizing the total log likelihood of $\sum_n \log p_\phi (y_n|\theta_n)$, which is approximately equivalent to minimizing
\begin{equation}
    \mathcal{L}_{\text{NL}}(\phi) = \E_{p(\boldsymbol{\theta})}(D_{\text{KL}} (p(y | \theta ) \| p_\phi (y | \theta) ) + \text{const},
    \label{eq:snl_loss}
\end{equation}
where the divergence is minimized when $p_\phi (y|\theta) \rightarrow p (y | \theta)$.

\subsection{Normalizing Flows}
\label{sec:flows}

Normalizing flows are flexible, tractable density estimators that model a target distribution via an invertible transformation of a simple base distribution \citep{Papamakarios2019, Kobyzev2019}. Let $u \sim p(u)$ (e.g., standard Gaussian) and let $f_\phi:\mathbb{R}^D \to \mathbb{R}^D$ be a bijection, typically a composition of $N$ invertible maps, $f_\phi = f_N \circ \cdots \circ f_1$. The induced density on $y=f_\phi(u)$ follows from change of variables,
\[
p_\phi(y) = p(u)\left|\det J_{f_\phi}(u)\right|^{-1},
\]
where $J_{f_\phi}$ is the Jacobian of $f_\phi$. Flows are trained by maximum likelihood on samples $\{y_n\}_{n=1}^N$, equivalently minimizing the forward KL divergence $D_{\mathrm{KL}}(p^*(y)\,\|\,p_\phi(y))$. In SBI, flows are commonly used in conditional form by conditioning the transformation (or its parameters) on context such as $\theta$ or $y$, yielding likelihood models $p_\phi(y\mid \theta,\xi)$ or posteriors $p_\phi(\theta\mid y)$ \citep{durka2019}. Because sampling is reparameterized (e.g., $y=f_\phi(u;\theta,\xi)$ with $u \sim p(u)$), flows admit pathwise gradients and are well-suited for gradient-based learning and variational objectives \citep{rezende2015variational, mohamed2019monte}.

\subsection{Bayesian Optimal Experimental Design}

\textbf{InfoNCE Bound} \hspace{0.5em} Following from Equation \ref{eq:infoGain}, \citet{ivanova2021implicit} proposed a lower bound of the MI based on the InfoNCE bound using an implicit likelihood with a differentiable simulator. They trained a design policy network $\pi_\psi$ that proposed designs based on the history of design-observation pairs, $h_{t-1} = \{(\xi_i, y_i)\}_{i=1:t-1}$ and a critic network $U_\phi$ that encapsulates the true likelihood when it maximizes the lower bound of MI. We adjust the bound to reflect the myopic experimental design setting (ignoring the design policy) and reformulate the bound as
\begin{equation}
    \label{eq:InfoNCE_objective}
    \mathcal{L}_\text{NCE}(\xi, \phi; L) \coloneqq \E \left[ \log \frac{\exp(U_\phi(y, \xi, \theta_0))} {\frac{1}{L+1}\sum_{i=0}^L \exp(U_\phi(y, \xi, \theta_i))} \right], 
\end{equation}
where the expectation is over ${p(\theta_0)p(y|\theta_0, \xi)p(\theta_{1:L})}$, $\xi$ is the proposed design, $\theta_0$ is the original parameter that generated data $y$, generated from the differentiable simulator $p(y|\theta_0, \xi)$, $L$ is the number of contrastive samples, and $U$ is a ``critic'' function such that $U\!:\!(y, \xi) \!\times\! \Theta \!\rightarrow\! \R$. This bound has low variance but is upper-bounded by $\log (L + 1)$, potentially leading to large bias with insufficient contrastive samples. Additionally, previous work required a differentiable simulator to take gradients with respect to the inputs of the simulator, which may not be available in SBI settings.

\subsection{Active learning in SBI}

SBI is often bottlenecked by the number of simulator evaluations, motivating \emph{active} strategies that prioritize which simulator queries to run. Sequential SBI methods such as SNPE/SNL already improve sample efficiency by adapting a proposal over parameters toward regions that better explain the observation, effectively steering simulations toward higher-posterior-density regions. However, proposal updates typically do not explicitly optimize the utility of individual simulator queries and may still allocate budget to redundant or low-value simulations. Bayesian active learning formalizes query selection through an acquisition function (often information-theoretic) that scores candidate parameters by their expected utility under the current model. A closely related recent approach, Active Sequential Neural Posterior Estimation (ASNPE) \cite{griesemer2024activesequentialposteriorestimation}, explicitly introduces an acquisition step into the SNPE loop by ranking candidate simulator parameters according to epistemic uncertainty in the current neural density estimator (approximated via Bayesian NDE machinery such as MC-dropout), and simulating only the top-ranked candidates. In our setting, this perspective motivates using prediction-oriented information-gain criteria (EPIG) to prioritize simulator calls when designs are fixed, yielding an active-learning view of SBI that complements BOED.

\section{SBI-BOED}
\label{sec:method}


We begin by noting that the mutual-information objectives used in BOED also provide a unifying view of several SBI procedures. In our setting, this shared MI perspective has two consequences. First, when experimental designs $\xi$ are optimized, maximizing an InfoNCE-based lower bound yields a practical objective for jointly training an amortized likelihood model and selecting informative designs, even for closed-box simulators. Second, when designs are fixed, the same learned likelihood can be used to prioritize simulator queries over latent parameters $\theta$, giving an active-learning strategy for allocating a limited simulation budget. We develop these two perspectives in turn: first through the connection between MI maximization and surrogate likelihood learning, and then through EPIG-based active learning of simulations.

\subsection{Bridging SBI and BOED}

We take inspiration from previous SBI and BOED methods to allow optimization of designs with respect to closed-box simulators that are modeled using normalizing flows. We start by noting how the loss function of contrastive ratio estimation (CRE) lower bounds \cref{eq:InfoNCE_objective} 
\begin{equation}
\begin{aligned}
    \log \frac{e^{g_\phi(\theta_0, y)}}{\frac{1}{L} \sum_{\ell=1}^L e^{g_\phi(\theta_{\ell}, y)}}
    &\leq \log \frac{e^{g_\phi(\theta_0, y)}}{\frac{1}{1 + L} \sum_{\ell=0}^L e^{g_\phi(\theta_{\ell}, y)}} \\[8pt]
    &= \log \frac{p_\phi(y | \theta_0, \xi)}{\frac{1}{1 + L}\sum^L_{\ell = 0} p_\phi(y | \theta_l, \xi)},
\end{aligned}
\end{equation}

where $L$ is the number of contrastive samples, which is $K$ in CRE, and $g_\phi$ is a discriminative classifier, which holds for a single batch of data and constant experimental design, i.e. when $\xi$ is constant. We use a neural density estimator to create a Likelihood-Free version of \cref{eq:InfoNCE_objective}. We now have a MI lower bound
\begin{equation}
    \mathcal{L}_{\text{NCE}}(\xi, \phi; L) \coloneqq \E \Bigg[\log \frac{ p_{\phi}(y | \theta_0, \xi)}{\frac{1}{1 + L}\sum^L_{\ell = 0} p_{\phi}(y | \theta_l, \xi)}  \Bigg]
    \label{eq:lf_pce}
\end{equation}
where the expectation is over $p(\theta_0)p(y|\theta_0, \xi)p(\theta_{1:L})$. We can now simultaneously optimize designs and parameters of a neural density estimator. If we are to use a normalizing flow instead of a classifier as $\exp g_\phi(y, \theta, \xi) = p_\phi(y|\theta, \xi)$, then the lower bound of the MI holds since normalizing flows are normalized probability distribution functions. The result is an amortized likelihood at a potentially optimal experimental design.  We discuss more SBI methods and their connection to MI optimization and BOED in \cref{sec:SBI+MI}.  Given this connection, we state our main proposition.

\begin{proposition}\label{thm:mi_max_sbi}
In the limit as the number of contrastive samples $L \to \infty$, maximizing the lower bound of the Mutual Information (MI) between parameters $\theta$ and observations $y$ in a Simulation-Based Inference (SBI) setting is equivalent, up to constants independent of $\phi$, to minimizing the Kullback-Leibler (KL) divergence, $D_{\text{KL}}(p(y | \boldsymbol{\theta}) || p_\phi(y | \boldsymbol{\theta}))$, between the likelihood $p(y | \boldsymbol{\theta})$ and its approximation $p_\phi(y | \boldsymbol{\theta})$, together with the corresponding marginal-likelihood approximation term, given as
\begin{align}
\begin{split}
    \max_{\phi} \mi_{\phi}(\theta ; y) \Leftrightarrow &\min_{\phi} \big[ \E_{p(\theta)} D_{KL}(p(y | \theta) || p_\phi(y | \theta)) \\
    &+ \E_{p_\phi(y)} D_{KL} (p(y) || p_\phi(y) ) \big],
\end{split}
\end{align}
where $\mi(\theta ; y)$ is the mutual information between $\theta$ and $y$, and $\mi_{\phi}(\theta ; y )$ is its approximation under parameter $\phi$.
\end{proposition}

Details and full proof of this proposition are in \cref{sec:MI_likelihood_proof}. This proposition characterizes the likelihood-learning interpretation of the MI objective. The same objective also has a complementary fixed-design interpretation: when $\xi$ is held constant, it can be used to score the value of simulator queries over $\theta$, yielding an active-learning strategy for simulation-efficient SBI.

\subsection{SBI Active Learning via EPIG}
\label{sec:parameter_active_learning}

While SBI-BOED focuses on selecting designs $\xi$ to maximize expected information gain, the same MI-based objective also induces an \emph{active learning} strategy over simulator parameters when designs are fixed (or when one wishes to prioritize which simulator queries to collect). Concretely, if $\xi$ is held constant, each simulator call corresponds to querying a parameter value $\theta$ and receiving an observation $y \sim p(y\mid \theta,\xi)$. Under a learned conditional likelihood $p_\phi(y\mid\theta,\xi)$, we can score candidate queries by the \emph{expected predictive information gain} (EPIG), which measures how strongly joint predictions under shared model uncertainty deviate from the product of marginals \cite{smith2024making, smith2023predictionorientedbayesianactivelearning}. Intuitively, EPIG prioritizes $\theta$ values for which the current likelihood model exhibits high epistemic uncertainty, yielding simulator calls that are expected to be most informative for improving the model.

To approximate epistemic uncertainty without requiring an explicit Bayesian neural network, we use MC-dropout to form a set of stochastic likelihood ``particles'' $\{\phi^{m}\}_{m=1}^M$. This is motivated by prior work showing that fixed-dropout-mask ensembles provide a practical mechanism for epistemic uncertainty estimation in normalizing flows \citep{berry2023normalizing}. For a candidate query $\theta$ and a set of target parameters $\theta^\star$ selected as the top-$K$ contrastive negatives under the InfoNCE objective, we estimate
\begin{equation}
\mathrm{EPIG}(\theta)\;\triangleq\;
\mathbb{E}\!\left[
\log \frac{p_\phi(y,y^\star \mid \theta,\theta^\star,\xi)}
{p_\phi(y\mid \theta,\xi)\,p_\phi(y^\star\mid \theta^\star,\xi)}
\right].
\label{eq:epig_theta}
\end{equation}
where the expectation is over $\,p(\theta^\star)\,p_\phi(y,y^\star \mid \theta,\theta^\star,\xi)$ and the joint predictive $p_\phi(y,y^\star \mid \theta,\theta^\star,\xi)$ is sampled by using the \emph{same} dropout particle $\phi^{(m)}$ for both $y$ and $y^\star$, capturing shared epistemic variation. We then acquire new simulator data by selecting $\theta$ with maximal EPIG and adding the resulting $(\theta,y)$ pairs to the training set. In this way, the same likelihood model used for BOED can be leveraged to make simulator calls more effective even in the absence of design optimization.

\subsection{Optimizing SBI-BOED}
\label{sec:optimizing_sbi_boed}

\textbf{Regularization} \hspace{0.5em} Stability of the density estimator is a challenge when optimizing the MI lower bound due to data distribution shift as a result of changing $p(y|\theta,\xi)$ when optimizing $\xi$. To address this, we added a regularization term, $\lambda$, to help stabilize the training of the density estimator during design optimization as
\begin{equation}
\begin{split}
    \mathcal{L}_{\text{NCE-}\lambda}(\xi, \phi; L) \coloneqq & \E \Bigg[\log \frac{ p_{\phi}(y | \theta_0, \xi)}{\frac{1}{1 + L}\sum^L_{\ell = 0} p_{\phi}(y | \theta_l, \xi)} \\
    &+ \lambda \cdot \log p_{\phi}(y | \theta_0, \xi)  \Bigg].
\end{split}    
\label{eq:lamb_lf_pce}
\end{equation}
where the expectation is over $p(\theta_0)p(y|\theta_0, \xi)p(\theta_{1:L})$. This regularization parameter gives more importance to accuracy of likelihood predictions when training where we provide theoretical analysis in \cref{sec:lambda_grads}.

\textbf{Regularized Mutual Information Estimator} \hspace{0.5em} We integrate the regularization term $\lambda$ into a mutual information estimator and propose the InfoNCE-$\lambda$ objective
\begin{equation} 
    I_{\text{NCE-}\lambda}(\phi, \lambda) \coloneqq \E_{p(\theta, y | \xi)} \left[ \log \frac{p_\phi(y|\theta, \xi)^{1+\lambda}}{\frac{1}{1 + L}\sum^L_{\ell = 0} p_{\phi}(y | \theta_l, \xi)} \right]. 
    \label{eq:info_nce_lambda}
\end{equation} 
This objective reweights the learned likelihood term through the exponent $1+\lambda$, allowing us to control the tradeoff between likelihood fitting and the resulting EIG estimate. In particular, varying $\lambda$ changes how strongly the objective emphasizes the surrogate likelihood relative to the marginal term. We provide a theoretical analysis of how $\lambda$ affects likelihood prediction accuracy and EIG values in \cref{sec:lambda_grads}.

\begin{algorithm}[tb]
\caption{SBI-BOED}
\label{alg:sbi-boed}
\begin{algorithmic}[1]
\STATE \textbf{Require:} Simulator $p(y|\theta, \xi)$, Estimator $p_\phi(y|\theta, \xi)$,\\
         Prior $p(\theta)$, Number of experimental designs $T$,\\
         Number of BOED training steps $N$
\STATE \textbf{Return:} Optimal designs $ \{\xi^*_i \mid i = 1, \ldots, T\} $,\\
        Approximate likelihood $p_{\phi}(y|\theta, \xi_T^*)$
\STATE Initialize dataset $\mathcal{D} = \{ \}$
\STATE Initialize $\hat{p}_{0}(\theta) \gets p(\theta)$
\FOR{$t=1,\ldots,T$}
    \STATE Sample $\theta_{0:L} \sim \hat{p}_{t-1}(\theta)$
    \FOR{$n=1,\ldots,N$}
        \STATE Sample $\xi \sim \mathcal{N}_{\text{trunc}}(\mu_\xi \mid \sigma_n^2)$
        \STATE Simulate $y \sim p(y|\theta, \xi)$
        \STATE Estimate $\nabla \mathcal{L}_{\text{NCE}-\lambda}(\xi, \phi; L)$ via \cref{eq:lamb_lf_pce}
        \STATE Update $\mu_\xi$ and $\phi$ using $\nabla_\xi$ and $\nabla_\phi$
        \STATE Checkpoint $\xi^* = \xi$ \textbf{if} $EIG_\xi > EIG_{\xi^*}$
    \ENDFOR
    \STATE Observe $y_o$ using $\xi^*$ in an experiment
    \STATE Add $(\xi^*, y_0)$ to dataset $\mathcal{D}$
    \STATE Set $\hat{p}_{t}(\theta) \propto \left( \prod_{(\xi_i, y_i) \in \mathcal{D}} p_\phi(y_i \mid \theta, \xi_i) \right) p(\theta)$
\ENDFOR
\end{algorithmic}
\end{algorithm}

\textbf{Design Distributions} \hspace{0.5em} A drawback of gradient-based methods is when there are sparse rewards, such as when the signal to noise ratio is, or approaches, zero. This results in designs struggling to optimize by gradient descent because optimizers may not be able to handle starting, or residing, in areas with no information.  We demonstrate this failing in an ablation study in Section \ref{sec:SIR_model} and take inspiration from RL's use of a replay buffer \citep{lin1992self, mnih2013playing} by imposing a parameterized distribution of designs and optimizing parameters of that distribution to return more information. When using design distributions, we compute the EIG for each individual observation $y_i$ associated with design $\xi_i$ and corresponding to the nominal parameter set $\theta_0$. In our usage, $\text{EIG}_i$ refers to the theoretical information gain expected from a single sample realization under design $\xi_i$, based on the current parameterization of the likelihood. It quantifies the information gain computed from a possible outcome as 
\begin{equation}
    \text{EIG}_i(\psi, \phi, L, \lambda) = \E \Bigg[\log \frac{ p_{\phi}(y_i | \theta_{0}, \xi_i)^{1 + \lambda}}{\frac{1}{1 + L}\sum^L_{\ell = 0} p_{\phi}(y_i | \theta_{l}, \xi_i)}\Bigg],
\end{equation}
with expectation over $p_\psi(\xi)p(\theta_0)p(y|\theta_0, \xi)p(\theta_{1:L})$ and where the parameter of the design distribution $\psi$ is optimized rather than the designs themselves.

We used a truncated Normal distribution for the design parameters, defined as $p_\psi(\xi) = \mathcal{N}_{\text{trunc}}(\mu_\xi| \sigma_n^2)$, where $\mathcal{N}_{\text{trunc}}$ denotes a Normal distribution truncated within the bounds $[\alpha, \beta]$, and $\alpha$ and $\beta$ are the lower and upper bounds, respectively, and $\sigma$ that uses a decays schedule depending on the training round, $n$. Whenever the rewards are sparse, if the design distribution is initialized with sufficient support then it will contain an EIG with significantly more reward that can be updated with gradient descent.  In our implementation, we used the reparameterization trick and a standard deviation schedule for $\sigma$ that decreased according to an exponential schedule $\sigma_n = \sigma_{\text{end}} + (\sigma_{\text{start}} - \sigma_{\text{end}}) \cdot e^{-\frac{n*\rho}{N}}$, where $\sigma_n$ is the new standard deviation for the Normal distribution that parameterizes $\xi$ in training round $n$, $\sigma_{start}$ and $\sigma_{end}$ are hyperparameters for the initial and final variances respectively, $N$ is the total number of rounds, and $\rho$ is a decay rate hyperparameter.

\textbf{Design Checkpoints} \hspace{0.5em} Checkpoints are used in supervised learning to save parameters that achieved low validation error \citep{vaswani2017attention, devlin2019bert}. \citet{fujimoto2023sale} proposed the use of checkpoints in RL applications to save a policy that obtains a high reward during training to help improve performance at test time. Even with a distribution of designs, we found designs falling into a local minima during optimization, such as in \cref{sec:design_dist_appendix}. We used design checkpoints, illustrated in \cref{fig:design_chkpt}, to mitigate the risk of gradient-based designs falling into local minima that do not contain sufficient support to encapsulate the global minimum loss. 

\textbf{Posterior Inference} \hspace{0.5em} The likelihood trained on the optimized design can return approximate posterior samples by sampling $\hat{p}(\theta | y, \xi) \propto p_\phi (y | \theta, \xi) p(\theta)$. Should the likelihood be trained on i.i.d. data, then the joint factorizes to product likelihood $\p{y_{1, \dots, N}|\theta, \xi_{1, \dots, N}} = \prod_{i=0}^N p(y_i|\theta, \xi_i)$ which can be used in Markov chain Monte Carlo (MCMC) to draw posterior samples.

\begin{figure*}[!t]
  \centering
  \includegraphics[width=\textwidth]{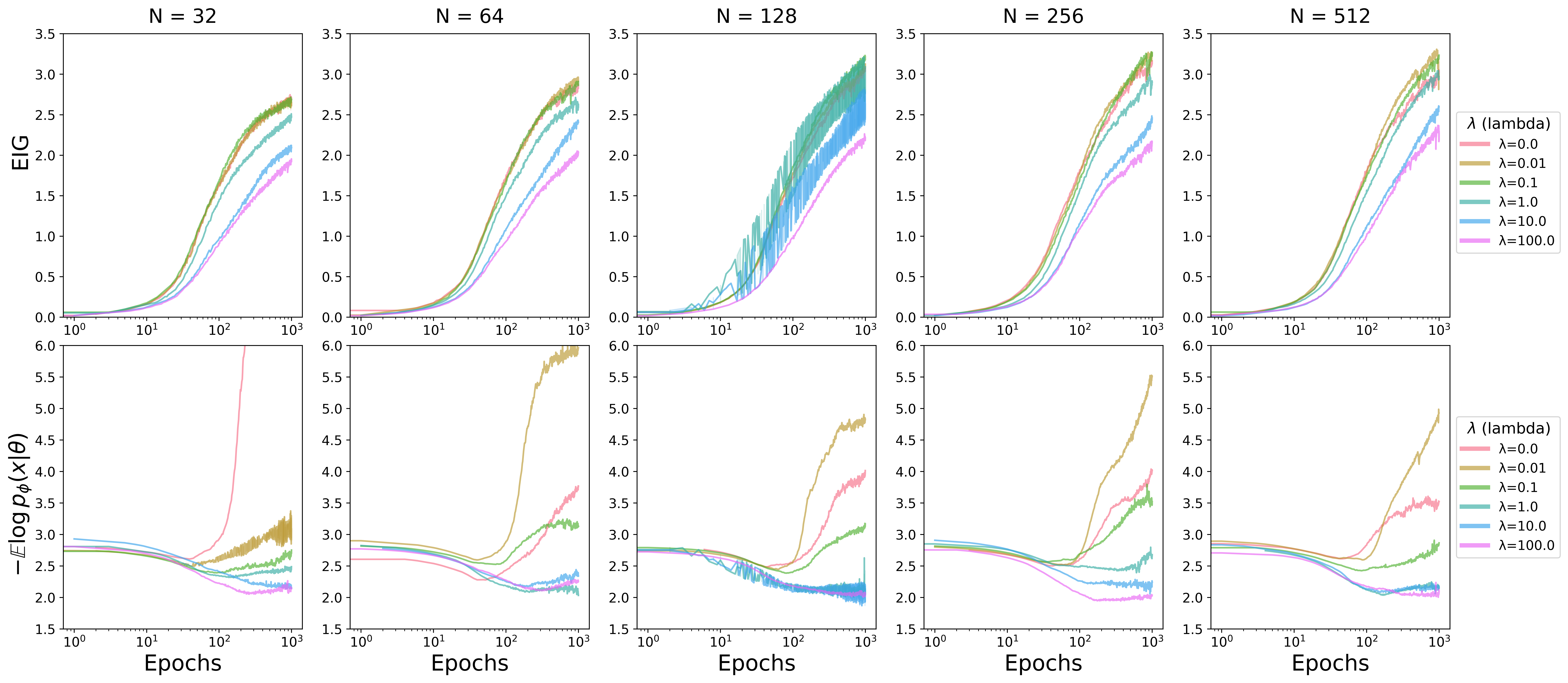}
  \vspace{-9pt}
  \caption{Comparison on the Two Moons task of the EIG and the validation loss $-\E \log p_\phi(y | \theta)$ across varying number of contrastive samples ($L = N - 1$) and $\lambda$ regularization. Increasing number of contrastive samples improves the information lower bound and likelihood validation metrics.  The $\lambda$ parameter helps improve the likelihood accuracy at the expense of MI.
  } 
  \label{fig:two_moons}
  \vspace{-12pt}
\end{figure*}

\section{Experiments}
\label{sec:experiments}



We evaluate the InfoNCE-$\lambda$ objective and its application to SBI-BOED across four settings. We first consider a fixed-design SBI benchmark (Two Moons), which isolates the optimization dynamics of the InfoNCE-$\lambda$ objective and enables the study of \emph{simulation active learning} via EPIG when $\xi$ is held constant (\cref{sec:parameter_active_learning}).

We then evaluate SBI-BOED on three design optimization tasks: (ii) a noisy linear model probing high-dimensional MI lower-bound optimization and the effect of $\lambda$ regularization; and (iii--iv) two sequential BOED problems, one time-dependent (SIR) and one i.i.d.\ (BMP). For BOED baselines, we compare against MINEBED-BO \citep{Kleinegesse2020b} and, when the simulator is differentiable, iDAD \citep{ivanova2021implicit} and MINEBED \citep{Kleinegesse2021}. We also include random and Sobol design SBI baselines to separate the value of design optimization from the likelihood objective (\cref{tab:sir_policy_objective_sweep,tab:bmp_policy_objective_sweep}).

For the BOED experiments (ii--iv), we report EIG as the primary information-gain metric and assess posterior quality via visual comparisons to NUTS samples and calibration metrics such as L-C2ST (with SBC in the appendix) \citep{linhart2023lc2stlocaldiagnosticsposterior}. Full experimental details are provided in \cref{sec:hyperparame}.

\subsection{Evaluation of MI on Two Moons}
\label{sec:two_moons}

We first study how well our amortized generative model performs in a \emph{fixed-design} (non-BOED) setting to isolate the behavior of the InfoNCE-$\lambda$ objective. This experiment highlights the tradeoffs between regularization in \cref{eq:lamb_lf_pce} and the number of contrastive samples $L$ in \cref{eq:InfoNCE_objective}. As noted by \citet{miller_contrastive_2022, glaser2022maximum}, aggressively optimizing MI between $y$ and $\theta$ can degrade likelihood fidelity, as reflected in the validation log probability $\mathbb{E}\,\log p_\phi(y\mid\theta)$. We therefore sweep $\lambda$ and $L$ and evaluate both information-gain proxies (EIG / MI lower bound) and likelihood quality (validation log probability). The resulting sweep is shown in \cref{fig:two_moons}: increasing $L$ improves the MI lower bound and generally improves likelihood validation metrics, while $\lambda$ controls the tradeoff between emphasizing the MI objective and likelihood accuracy. We note that, with our chosen parameterization, optimizing \cref{eq:InfoNCE_objective} does not eliminate the known mode collapse behavior in Two Moons likelihood-based models \citep{greenberg2019automatic}; an example is shown in \cref{sec:two_moons_reloaded}.

\paragraph{InfoNCE-$\lambda$ active learning}In addition to this diagnostic sweep, Two Moons provides a simple setting to evaluate \emph{active learning of simulations} when $\xi$ is fixed: rather than drawing simulator queries $\theta \sim p(\theta)$ i.i.d., we select $\theta$ using EPIG computed from the current learned likelihood with MC-dropout (\cref{sec:parameter_active_learning}). We track calibration over sequential rounds using C2ST trajectories and summarize final local calibration using the L-C2ST statistic. Across this sweep, $\lambda$ has the most pronounced effect on calibration: smaller $\lambda$ yields better C2ST trajectories and lower final L-C2ST values (better calibration), while larger $\lambda$ tends to worsen local calibration (Fig.~\ref{fig:two_moons_c2st_lc2st}). The contrastive top-$K$ has a less systematic, non-monotone effect and primarily acts as a variance-control knob; increasing $K$ is relatively inexpensive (increase bank size) but needs to be balanced with the number of $\theta$ samples. Finally, we compare EPIG-based querying to an i.i.d.\ (random) $\theta$ baseline and observe that EPIG yields more efficient improvement in calibration over rounds under the same simulation budget (details in \cref{sec:two_moons_active_appendix}).

\subsection{Noisy Linear Model}

In our first BOED task, we evaluate how SBI-BOED performs on the EIG metric with increasing design dimensions. We follow \citet{Kleinegesse2020b} and evaluate optimal designs on a noisy linear model where a response variable $y$ has a linear relationship with experimental designs $\xi$, which is determined by values of the model parameters $\boldsymbol{\theta} = [\theta_0, \theta_1]$, which model the offset and gradient. We would like to optimize the value of $D$ measurements to estimate the posterior of $\boldsymbol{\theta}$, and so create a design vector $\boldsymbol{\xi} = [\xi_1, \dots, \xi_D]^{\mathsf{T}}$.  Each design, $\xi_i$ returns a measurement $y_i$, which results in the data vector $\boldsymbol{y} = [y_1, \dots, y_D]^{\mathsf{T}}$. We use a Gaussian noise source $\mathcal{N}(\epsilon; 0, 1)$ and Gamma noise source $\Gamma (\nu; 2, 2)$. The model is then 
\begin{equation}
    \boldsymbol{y} = \theta_0\boldsymbol{1} + \theta_1 * \boldsymbol{\xi} + \boldsymbol{\epsilon} + \boldsymbol{\nu},
\end{equation}
where $\boldsymbol{\epsilon} = [\epsilon_1, \dots, \epsilon_D]^{\mathsf{T}}$ and $\boldsymbol{\nu} = [\nu_1, \dots, \nu_D]^{\mathsf{T}}$ are i.i.d. samples. We used a prior distribution on model parameters as $p(\boldsymbol{\theta}) = \mathcal{N}(\boldsymbol{\theta} ; 0, 3^2)$.  We evaluate SBI-BOED, \textit{purely using design gradients and no distribution over designs}, to examine how changing the $\lambda$ regularization parameter in \cref{eq:lamb_lf_pce} influences the resulting MI bound and design optimization stability with increasing design dimension.

For all design dimensions, we randomly initialize designs $\xi \in [-10, 10]$. For SBI-BOED, we chose $N=10$, the number of batch samples $\boldsymbol{y} \sim p(\boldsymbol{y}|\theta_0, \boldsymbol{\xi})$, and $L=50$ contrastive samples. We used a neural spline flow with training details in \cref{sec:hyperparame}.
We show the plots of the EIG in \cref{fig:linear_eig}, where we can see that SBI-BOED optimizes a lower bound of MI, which increases for higher design dimensions. This corresponds with intuition that gathering more data results in more information. Similar to the two moons task, we see how the choice of $\lambda$ influences the stability of MI estimation for SBI-BOED in high design dimensions where there seems to be a tradeoff with choice of regularization and MI estimation. This may be due to regularization limiting gradient updates of designs that lead to out of distribution data distributions, but is balanced by the stability in the training objective.

\subsection{SIR Model}
\label{sec:SIR_model}

\begin{table*}[h]
\centering
\small
\setlength{\tabcolsep}{3pt}
\renewcommand{\arraystretch}{1.1}
\caption{SIR (Section \ref{sec:SIR_model}) and BMP (Section \ref{sec:BMP_model}) results: comparison of EIG, L-C2ST, and median distance (Med.) for two tasks with $T=2$ experimental design rounds for the SIR model and $T=3$ for the BMP model. For the InfoNCE-$\lambda$ baselines, we report the best $\lambda$ by EIG for each task and design policy. We report mean and standard error over 3 experiments using different seeds. On BMP, wall-clock time was $2,235 \pm 9$ s for SBI-BOED and $7,095 \pm 2,519$ s for MINEBED-BO under the matched-budget sweep.}
\vspace{-6pt}
\label{tab:exp_compare_expanded}
\begin{tabular*}{\textwidth}{@{\extracolsep{\fill}}l*{6}{c}}
    \toprule
    \multirow{2}{*}{\textbf{Method}} & \multicolumn{3}{c}{\textbf{SIR (T=2)}} & \multicolumn{3}{c}{\textbf{BMP (T=3)}} \\
    \cmidrule(lr){2-4} \cmidrule(lr){5-7}
    & EIG $(\uparrow)$ & L-C2ST$^\dagger$ $(\downarrow)$ & Med. $(\downarrow)$ & EIG $(\uparrow)$ & L-C2ST$^\dagger$ $(\downarrow)$ & Med. $(\downarrow)$ \\
    \midrule
    Random + NLE                        & 0.02 $\pm$ 0.02 & 0.01 $\pm$ 0.01 & 78.97 $\pm$ 33.23 & 0.02 $\pm$ 0.01 & 0.01 $\pm$ 0.01 & 11.49 $\pm$ 5.33 \\
    Random + InfoNCE-$\lambda$ ($\lambda = 0.01$)    & 0.93 $\pm$ 0.36 & 0.01 $\pm$ 0.01 & 77.82 $\pm$ 33.09 & 0.46 $\pm$ 0.07 & 0.01 $\pm$ 0.01 & 10.86 $\pm$ 5.72 \\
    \addlinespace[2pt]
    \cmidrule(lr){1-7}
    \addlinespace[2pt]
    MINEBED(-BO)                        & \textbf{2.69 $\pm$ 0.03} & 0.07 $\pm$ 0.05 & 71.26 $\pm$ 5.66 & \textbf{14.23 $\pm$ 0.60} & 0.25 $\pm$ 0.00 & 0.82 $\pm$ 0.06 \\
    iDAD (InfoNCE)                      & 2.67 $\pm$ 0.02 & 0.10 $\pm$ 0.04 & 71.65 $\pm$ 2.78 & - & - & - \\
    SBI-BOED ($\lambda = 1$)            & 1.01 $\pm$ 0.03 & 0.05 $\pm$ 0.01 & 47.99 $\pm$ 2.62 & 10.39 $\pm$ 0.01 & 0.01 $\pm$ 0.01 & 0.61 $\pm$ 0.01 \\
    SBI-BOED ($\lambda = 0.1$)          & 1.47 $\pm$ 0.23 & 0.03 $\pm$ 0.01 & \textbf{46.85 $\pm$ 3.21} & 10.38 $\pm$ 0.01 & 0.01 $\pm$ 0.01 & \textbf{0.60 $\pm$ 0.01} \\
    SBI-BOED ($\lambda = 0.01$)         & 1.63 $\pm$ 0.23 & 0.04 $\pm$ 0.02 & 52.85 $\pm$ 0.70 & 10.39 $\pm$ 0.01 & 0.01 $\pm$ 0.01 & 0.61 $\pm$ 0.01 \\
    \bottomrule
\end{tabular*}
\vspace{2pt}
\begin{flushleft}
\footnotesize
$^\dagger$ L-C2ST values smaller than $0.01$ are displayed as $0.01$ for readability.
\end{flushleft}
\vspace{-9pt}
\end{table*}

\begin{figure}[h] 
  \centering
  \vspace{-3pt}
  \includegraphics[width=\columnwidth]{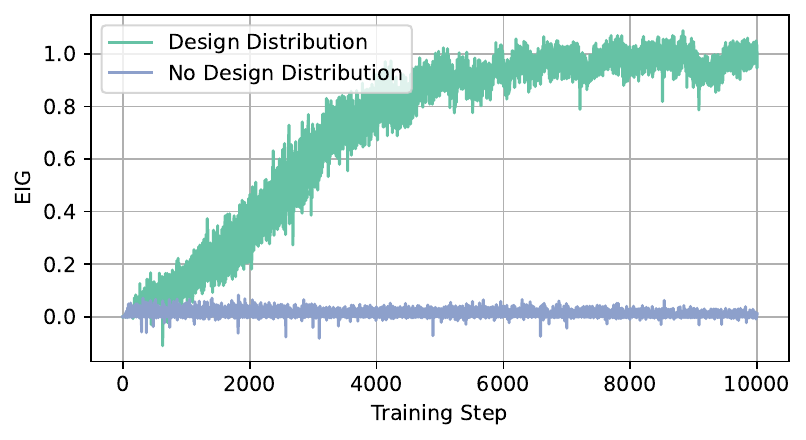}
  \vspace{-23pt}
  \caption{Training (higher is better) without a design distribution in the first round of optimization for the SIR model fails to find designs with high rewards.}
  \label{fig:ablation_study}
  \vspace{-6pt}
\end{figure}

We next evaluate the performance of SBI-BOED with different amounts of regularization on a real-world implicit likelihood model that has a differentiable simulator as a comparison to alternative methods. We use the Susceptible, Infected, or Recovered (SIR) epidemiology model specified in \citet{ivanova2021implicit}. The SIR model represents a fixed population that has three groups: susceptible, infected, and recovered. Individuals transition from susceptible to infected with a parameter $\beta$ and from infected to recovered with parameter $\gamma$. Therefore, the model parameters are $\theta \coloneqq [\beta, \gamma]$ and the design space $\Xi$, consists of a time $\kappa$ to measure individuals to infer parameters $\theta$. We measured the EIG, L-C2ST, and median distance from the observed data point to samples generated by the final posterior distribution with ground truth parameters $\theta = [0.7399, 0.0924]$. \cref{tab:exp_compare_expanded} summarizes the results. We find that iDAD and differentiable MINEBED achieve the best information gain, likely thanks to using differentiability of the simulator, but perform worse than SBI-BOED on L-C2ST and median distance metrics. This may indicate the learned critic is not as accurate as the amortized likelihood trained in SBI-BOED. Among SBI-BOED methods, we see that more regularization generally leads to improved median distance, and all perform roughly the same on calibration. Qualitative predictive, posterior, and SBC diagnostics for a representative SIR posterior are shown in \cref{fig:sir_posterior}. Thus, improved information gain \textit{does not} necessarily correlate with improved downstream prediction, as measured by the median distance metric. This highlights an important caveat to BOED methods that we should not rely on a single metric and holistically assess the resulting inference model in terms of accuracy and calibration before making decisions (performing experiments). 

\textbf{Ablation Study} \hspace{0.5em} We compare the performance of SBI-BOED with and without a design distribution when maximizing the EIG in \cref{fig:ablation_study} in the first round of design optimization. The prior used in the SIR model creates a challenge for myopic and local design optimization by introducing regions with little EIG signal, such as the flatter part of the SIR curve shown in \cref{sec:boed_reloaded}. The design distribution overcomes this challenge by querying diverse sets of designs with better gradients to optimize \cref{eq:lamb_lf_pce}.

\subsection{Bone Morphogenetic Protein Model}
\label{sec:BMP_model}

The Bone Morphogenetic Protein (BMP) pathway is important in developmental and disease processes. A mass action kinetics model was proposed for the BMP pathway \citep{antebi2017combinatorial}. The one-step model proposed by \citet{su2022ligand} models type I (A) and type II (B) receptors coupling with a ligand (L) to form a trimer complex (T) with equilibrium affinity $K$ in a single step as $A + B + L \autorightleftharpoons{$K$}{} T$. The trimeric complex then phosphorylates SMAD protein to send a downstream gene expression signal, $S$, with a certain efficiency, $\varepsilon$ as $\varepsilon T = S$. Steady-state gene expression signals can be simulated using convex optimization in a closed-box optimization process. Thus, we would like to infer the parameters $K$ and $\varepsilon$  given data, $S$. In the experimental design context, we would like to optimize the ligand concentration, $\boldsymbol{L}^*$, used in an experiment so to gain the most information about the parameters of this model of a biological signaling pathway. The model parameters are $\theta \coloneqq [K, \varepsilon]$ and we use ground truth parameters $\theta = [0.85, 0.85]$. We show qualitative predictive, posterior, and SBC diagnostics for BMP in \cref{fig:bmp_posterior}. We evaluated a 1D design dimension for the ligand concentration on a uniform range from $10^{-3}\text{ng/mL}$ to $10^{3}\text{ng/mL}$. We compare against the same benchmarks except the iDAD algorithm, which cannot work on this non-differentiable simulator. We train the EIG for 500 steps in each design optimization round because of the expensive simulation time required in each round (about 15 seconds). All SBI-BOED design algorithms perform well in this setting, partially due to the bias in larger concentrations containing more information. Under the matched-budget MINEBED-BO comparison, MINEBED-BO obtains higher estimated EIG than SBI-BOED ($14.23 \pm 0.60$ vs.\ $10.39 \pm 0.01$), but SBI-BOED achieves better posterior metrics, with lower L-C2ST ($0.01 \pm 0.01$ vs.\ $0.25 \pm 0.00$) and lower median distance ($0.60 \pm 0.01$ vs.\ $0.82 \pm 0.06$). Thus, the model with the best EIG does not correspond to the most-calibrated model nor the best accuracy predictions. 

\section{Discussion}
\label{sec:discussion}

This work connects BOED-style mutual-information maximization with likelihood learning in SBI by showing that an InfoNCE-based MI lower bound with a normalized conditional density model contains a likelihood-fitting term and a finite-contrastive marginal-likelihood approximation while optimizing for informative queries. Empirically, our Two Moons study illustrates the central tradeoff induced by the InfoNCE-$\lambda$ objective: regularization and contrastive sampling affect not only MI estimates (EIG) but also likelihood fidelity and posterior calibration, and EPIG-based simulation active learning can improve calibration over sequential rounds under a fixed simulation budget. In sequential BOED benchmarks, we further observe that designs with higher EIG do not necessarily yield better downstream inference, such as posterior calibration and predictive accuracy diverging from information-gain objectives, and highlighting the need to evaluate BOED methods beyond EIG alone.

On the optimization side, we introduced practical stabilizers for closed-box simulators and purely generative surrogates, including optimizing a distribution over designs to mitigate sparse-reward landscapes and design checkpointing to avoid local optima during gradient-based search. Across differentiable and non-differentiable simulators, SBI-BOED achieves competitive or improved information gain while consistently improving calibration and predictive metrics relative to strong BOED baselines, supporting the view that simulation efficiency should be measured by the quality of the resulting posterior, not solely by utility values.

Finally, while we instantiate SBI-BOED with conditional normalizing flows, the approach extends to any likelihood-based or likelihood-bounded generative model that supports evaluating (or lower-bounding) $\log p_\phi(y\mid\theta,\xi)$. This suggests natural extensions to more expressive generative families such as diffusion models and flow-matching \citep{ho2020denoising, song2020generativemodelingestimatinggradients, lipman2023flow, pmlr-v235-ben-hamu24a}, which may better handle high-dimensional observations and offer additional flexibility for design optimization through alternative parameterizations of the data and noise spaces. However, these approaches add complexity over our simple approach that should be a starting point for scientific simulators suited to likelihood-based SBI modeling.

\bibliography{uai2026-template}

\appendix
\onecolumn

\section{Mutual Information-Based Likelihood Optimization Derivation \& Proofs}
\label{sec:MI_proofs}

We provide a derivation and proof of how optimizing a lower bound of the mutual information is the same as minimizing the KL divergence in \eqref{eq:snl_loss}. We also discuss alternative BOED bounds and how adding a distribution of designs to the MI approximation keeps it a valid lower bound.

\subsection{Derivation of Optimization of the Mutual Information}
\label{sec:MI_likelihood_proof}

We investigate how optimizing the mutual information within the SBI-BOED loss framework leads to an implicit optimization of the likelihood. Following the approach outlined by \cite{miller_contrastive_2022}, we optimize our approximation to the true likelihood by minimizing the KL divergence:

\begin{equation}
    D_{\text{KL}}(p(y | \theta) || p_\phi(y | \theta)).
\end{equation}

Within the SBI framework, we draw samples of parameters \( \theta \) from the prior \( p(\theta) \) and of data \( y \) conditioned on these parameters from the likelihood \( p(y | \theta) \). This sampling enables us to approximate the expected KL divergence across the parameter space:

\begin{equation}
    \mathbb{E}_{p(\theta)}[D_{\text{KL}}(p(y | \theta) || p_\phi(y | \theta))].
\end{equation}

Now we express the KL divergence as the expectation of the log ratio of probabilities:

\begin{align}
    \mathbb{E}_{p(\theta)}[D_{\text{KL}}(p(y | \theta) || p_\phi(y | \theta))] &= \mathbb{E}_{p(\theta, y)} \left[ \log \frac{p(y|\theta)}{p_\phi(y | \theta)} \right] \\
    &= \mathbb{E}_{p(\theta, y)} \left[ \log \frac{p(y|\theta)}{p(y)} \frac{p(y)}{p_\phi (y|\theta)} \right] \\ 
    &= \mi(\theta ; y) + \mathbb{E}_{p(\theta, y)} \left[ \log \frac{p(y)}{p_\phi(y|\theta)} \right] \\
    &= \mi(\theta ; y) + \mathbb{E}_{p(y)} \left[ \log p(y) \right] - \mathbb{E}_{p(\theta, y)} \left[ \log p_\phi (y | \theta) \right].
\end{align}

Since the KL divergence is always non-negative, the optimization process aims to find the parameters \( \phi \) that minimize this expected divergence, implicitly maximizing the mutual information between parameters.

We now base our proof of \cref{thm:mi_max_sbi} as follows:

\begin{proof}
    Let us consider the MI between random variables $\Theta$ and $Y$, where $y \sim Y$ and $\theta \sim \Theta$, then we have

    \begin{equation}
        \mi(\theta ; y) = \mathbb{E}_{p(\theta, y)} \left[ \log \frac{p(y|\theta)}{p(y)} \right].
    \end{equation}

    In the SBI setting, we would like to approximate the likelihood $p_\phi(y|\theta)$. We can approximate the marginal likelihood by the Strong Law of Large Numbers using $\E_{p(\theta)} \big[ p(y|\theta) \big] = \int_\theta p(y|\theta) p(\theta) d\theta \approx \frac{1}{L} \sum_\ell^L p(y|\theta_\ell )$. The bound gets tighter as $L\rightarrow \infty$ and as the likelihood better-approximates the true likelihood. Assuming we are optimizing the parameters $\phi$ to maximize this objective, then we have

    \begin{align}
        \hat{\phi} &= \underset{\phi}{\arg\max} \; \E_{p(\theta, y)} \left[ \log \frac{p_\phi(y|\theta)}{\hat{p}(y)} \right] \\
        &= \underset{\phi}{\arg\min} \; \left( I(\theta; y) - \E_{p(\theta, y)} \left[ \log \frac{p_\phi(y|\theta)}{\hat{p}(y)} \right] \right) \\
        &= \underset{\phi}{\arg\min} \; \E_{p(\theta, y)} \left[ \log \frac{p(y|\theta)}{p(y)} \frac{\hat{p}(y)}{p_\phi(y|\theta)} \right] \label{lambda_intro_eq} \\
        &\approx \underset{\phi}{\arg\min} \; \left( \E_{p(\theta)}D_{KL}(p(y | \theta) || p_\phi(y | \theta)) - \E_{\hat{p}(y)}D_{KL}(\hat{p}(y) || p(y)) \right) \\
        &= \underset{\phi}{\arg\min} \left( \E_{p(\theta)} D_{KL}(p(y | \theta) || p_\phi(y | \theta)) + \E_{\hat{p}(y)} \left[ \log \hat{p}(y) \right] - \text{const.}\right) \label{mi_kl_combo}
    \end{align}

    where the marginal likelihood is approximated as $\hat{p}(y) = \frac{1}{L}\sum_{i=1}^L p_\phi(y|\theta_i), \quad \theta_i \sim p(\theta)$. Thus, as $L \to \infty$, maximizing the InfoNCE bound from \cref{eq:InfoNCE_objective} is asymptotically equivalent to optimizing the likelihood objective in \cref{eq:snl_loss}, with a finite-$L$ marginal-likelihood approximation term whose accuracy depends on the number of contrastive samples $L$.

\end{proof}

\textit{Remark.} In the BOED setting, the MI is simply conditional on a design, $\xi$, as $\mi(\theta;y|\xi)$ which allows for gradient-based optimization by a pathwise gradient estimator as detailed in \cref{sec:flows}. An interesting insight to this derivation is that InfoNCE bound used to maximize a lower bound of MI may be biased by the reverse KL of the marginal likelihood to models that better represent the data. This may surface in BOED with designs preferring models that do not cover enough of the potential parameter spaces that explain the data.

\section{Theoretical Analysis of SBI-BOED Components}

We provide more theoretical analysis of the impact of the $\lambda$ parameter of $I_{\text{NCE-}\lambda}$ on predictive accuracy, estimating the EIG, and gradients of EIG and normalizing flow model parameters, $\phi$. 

\subsection{Analysis of the \texorpdfstring{$\lambda$}{lambda} regularization parameter}
\label{sec:lambda_grads}

We experimentally demonstrated in \cref{sec:two_moons} that the $\lambda$ regularization parameter in SBI-BOED influences both the likelihood's validation accuracy and the EIG bound. We theoretically analyze both scenarios as well as $\lambda$ influence on optimization gradients in the following sections.

\textbf{Influence of $\lambda$ on likelihood prediction accuracy} \hspace{0.5em} Starting from \cref{lambda_intro_eq} of the previous section, including the $\lambda$ regularization parameter results in
\begin{align}
    \hat{\phi} &= \underset{\phi}{\arg\min} \; \E_{p(\theta, y)} \left[ \log \frac{p(y|\theta)}{p(y)} \frac{\hat{p}(y)}{p_\phi(y|\theta)} \right] + \lambda \E_{p(\theta, y)} \log [p_\phi(y | \theta)] \\
    &= \underset{\phi}{\arg\min} \; \E_{p(\theta, y)} \left[ \log \frac{p(y|\theta)}{p(y)} \frac{\hat{p}(y)}{p_\phi(y|\theta)} p_\phi(y | \theta)^{\lambda} \right].
\end{align}

We now focus on the contribution of the added $p_\phi(y | \theta)^{\lambda}$ term grouped with the likelihood's KL divergence from \cref{mi_kl_combo}. Reformulating this term, we express the $\lambda$-regularized KL divergence as
\begin{align}
    D_{\text{KL}}(p(y|\theta) \| p_\phi(y|\theta)^{1-\lambda}) &= \mathbb{E}_{p(y|\theta)} \left[ \log \frac{p(y|\theta)}{p_\phi(y|\theta)^{1-\lambda}} \right] \\
    &= \mathbb{E}_{p(y|\theta)} \left[ \log p(y|\theta) - (1-\lambda) \log p_\phi(y|\theta) \right].
    \label{eq:gradient_phi}
\end{align}

This modified divergence adjusts the weighting of the log-likelihood term \(\log p_\phi(y|\theta)\) based on \(\lambda\). We show how $\lambda$ influences optimization, omitting the trivial case when $\lambda = 0$:
\begin{itemize}
    \item For $\lambda > 0$: The term $(1-\lambda)$ reduces the weight of the approximate likelihood, ensuring broad coverage of the parameter space while emphasizing accuracy.
    \item For $\lambda < 0$: The term $(1-\lambda) > 1$ amplifies the weight of the approximate likelihood, leading to mode-seeking behavior that prioritizes high-probability regions at the expense of the tails.
\end{itemize}


\textbf{Influence of $\lambda$ on the EIG} \hspace{0.5em} We start from \cref{eq:info_nce_lambda} and rephrase it here using the shortened marginal likelihood notation $\hat{p}(y | \xi)$ for convenience
\begin{align}
    \E_{p(\theta, y | \xi)} \left[ \log \frac{p_\phi(y|\theta, \xi)^{1+\lambda}}{\hat{p}(y|\xi)} \right].
\end{align}

Expanding the log ratio, we can separate the contributions of the likelihood and the marginal likelihood
\begin{align}
    \E_{p(\theta, y | \xi)} \left[ \log \frac{p_\phi(y|\theta, \xi)^{1+\lambda}}{\hat{p}(y|\xi)} \right] = \E_{p(\theta, y | \xi)} \left[ (1+\lambda) \log p_\phi(y|\theta, \xi) - \log \hat{p}(y|\xi) \right].
\end{align}

The parameter $\lambda$ influences the scaling of the likelihood term $(1+\lambda) \log p_\phi(y|\theta, \xi)$ and indirectly affects the marginal likelihood $\hat{p}(y|\xi)$ through its dependence on $p_\phi(y|\theta, \xi)$. The EIG linearly depends on the $\lambda$ value. Positive $\lambda$ decrease the EIG estimate while negative values increase the EIG at the expense of likelihood approximation, as previously discussed. The decrease in EIG with increasing $\lambda$ aligns with empirical observations (\cref{fig:two_moons}). 

\section{Related Work}
\label{sec:related_work}

In BOED, we focus on the setting of myopic gradient-based experimental design. Previously, \citet{Foster2019} proposed various likelihood-free information bounds that could work in the SBI setting based on variational inference estimators but did not make use of a normalized generative model like a normalizing flow. This is problematic for use in sequential SBI methods that make use of a normalized likelihood or posterior functions. \citet{Kleinegesse2020b} developed MINEBED to simultaneously optimize experimental designs and a critic that could draw samples from the posterior, but relied on a differentiable simulator or using Bayesian optimization. \citet{ivanova2021implicit} extended both previous works and developed policy-based experimental designs for non-myopic experimental designs, but whose critics also relied on differentiable simulators - simulators whose inputs can be connected to a differentiable computation graph, which may not be available for scientific simulators. An RL approach \citep{lim2022policybased} optimized a critic without requiring a differentiable simulator but relies on computationally expensive RL algorithms and may not be feasible for practitioners with scientific simulators that can take a significant amount of time to simulate. This is exemplified in our biological experiment where each round of simulation takes about ten seconds.

Recent posterior- and variational-BOED methods also use normalizing flows or amortized posteriors to optimize design objectives \citep{dong2025variational,orozco2024probabilistic,shen2025variational}. These approaches are complementary to SBI-BOED, but they largely target a different operating regime. \citet{dong2025variational} can handle implicit likelihoods through sampling, but simultaneous design optimization relies on gradients through the design path or an equivalent differentiable construction. \citet{orozco2024probabilistic} similarly uses conditional normalizing flows with differentiable simulator/design structure. \citet{shen2025variational} addresses sequential design with reinforcement learning, but shifts the cost to large numbers of simulated rollouts. In contrast, our evaluation focuses on expensive closed-box simulators where simulator/design gradients are unavailable, simulator calls are budget-limiting, and the learned likelihood should remain reusable for posterior inference after design optimization.

Recent work connecting SBI methods to MI-based optimization studied how to stably train a discriminative and generative SBI model \citep{miller2023simulation}, and applying a nonparametric function to the selection of data points to reduce variance of the resulting MI estimate \citep{glaser2022maximum}. These methods take a complementary approach to ours but require a generative and discriminative model whereas we only require a generative model. Additionally, we study the utility of MI-based optimization of SBI models in BOED.

\section{Implementation of SBI-BOED in Posterior Estimation, Ratio estimation, and checkpointing}
\label{sec:SBI+MI}

\subsection{Applying the InfoNCE Bound to Neural Posterior Estimation}
\label{sec:snpe_mi}

We demonstrate the theoretical basis for optimizing a Neural Posterior Estimation (NPE) network and experimental designs, which learns a surrogate posterior conditioned on simulated data, $p_\phi(\theta | y, \xi)$. 

\textbf{Neural Posterior Estimation} \hspace{0.5em} Previous methods for directly estimating the posterior by maximum likelihood estimation struggled with bias \cite{Papamakarios2016} or variance \cite{lueckmann2017flexible}. An alternative method was developed by \citet{greenberg2019automatic} to recover the unbiased posterior,
\begin{equation}
    p(\theta | y) \approx \frac{ p_\phi (\theta | y) / p(\theta) }{ Z_\phi (y)}
\end{equation}
by calculating the normalizing constant $Z_\phi = \int p_\phi (\theta | y)/p(\theta) d\theta $. They then minimize $\mathcal{L}_{\text{NP}}(\phi) = \E_{p(y|\theta)p(\theta)}[-\log p_\phi (\theta | y)]$ to return an amortized posterior. Except for mixture density networks \citep{bishop1994mixture}, the normalizing constant is difficult to approximate. \citet{greenberg2019automatic} alternatively proposed to approximate the intractable normalizing constant by replacing the integral with a summation term that takes draws of the parameters from a proposal set $\Theta$ such that the posterior is approximately:
\begin{equation}
    p(\theta | y) \approx \frac{ p_\phi (\theta | y) / p(\theta) }{ \sum_{\theta' \in \Theta} p_\phi (\theta' | y) / p(\theta')}.
    \label{eq:snpe}
\end{equation}


\textbf{Applying NPE in BOED} \hspace{0.5em} We can use this in experimental design by again using the reparameterization trick to pass gradients back to the conditional $\xi$ inputs to the approximate posterior. In some cases, it may be more desirable to directly infer a posterior distribution instead of a likelihood. For example, if the data, $y_o$, to train a flow is computationally infeasible but whose latent parameters, $\theta$, is sufficiently small for use in a normalizing flow. Since the posterior is proportional to the likelihood, $p(\theta | y) \propto p(y|\theta)p(\theta)$, we can replace the likelihood in \cref{eq:lf_pce} with a posterior
\begin{equation}
    \label{eq:NP_InfoNCE}
    \mathcal{L}_{\text{NCE-NPE}}(\xi, \phi, L) \coloneqq \E_{p(\theta_0)p(y|\theta_0, \xi)p(\theta_{1:L})}  \frac{ p_\phi (\theta_0 | y, \xi) / p(\theta_0) }{ \frac{1}{1 + L}\sum_{\ell = 0}^L p_\phi (\theta_\ell | y, \xi) / p(\theta_\ell)},
\end{equation}
which is a biased lower bound of the MI by a factor of the prior $p(\theta)$, but whose gradient can still be used to train a density estimator and optimize experimental designs. This case should be used when it is desirable to have an amortized posterior and the absolute EIG does not matter.

\subsection{The NWJ BOED Bound \& Contrastive Ratio Estimation}
\label{sec:nwj_bound}

We demonstrate the theoretical basis for optimizing Neural Ratio Estimation (NRE) network and experimental designs, which learns a surrogate ratio estimator conditioned on experimental designs, $g_\phi(y,\theta|\xi)$. 

\textbf{Neural Ratio Estimation} \hspace{0.5em} Classifiers can be used to approximate the likelihood-to-evidence ratio such that $g_\phi (\theta, y) \approx \log \frac{p(y|\theta)}{p(y)} + c(y)$, where $c(y)$ is a bias term introduced from approximating the ratio, and which can be minimized \citep{Durkan2020, hastie2009elements}. The classifier can be trained by minimizing the loss 
\begin{equation}
    \mathcal{L}_{\text{CRE}}(\phi) = - \frac{1}{B} \sum_{b=1}^B \log \frac{\exp(g_\phi (\theta^{(b)}, y^{(b)} )) }{\sum_{k=1}^K \exp (g_\phi (\theta^{(k)}, y^{(b)} )) }
    \label{eq:nre}
\end{equation}

over $B$ batches of contrasting parameters. Given the use of contrastive samples, \citet{Durkan2020} called this contrastive ratio estimation (CRE) and noted the similarity between CRE and the Noise Contrastive Estimation (InfoNCE) MI lower bound  proposed by \citet{poole2019variational}.

\textbf{Applying NRE in BOED} \hspace{0.5em} We present another relevant MI bounds to this paper, the NWJ \cite{Nguyen_2010} bound has been used in BOED in \citet{Kleinegesse2021, ivanova2021implicit}.  Adjusting the bound from \citet{ivanova2021implicit},
\begin{equation}
    \label{eq:NWJ_objective}
    \mathcal{L}_\text{NWJ} (\xi, \phi) \coloneqq \E_{p(\theta)p(y|\theta, \xi)} \left[ g_\phi(y, \theta) \right] - e^{-1} \E_{p(\theta)p(y | \xi)} \left[ \text{exp}(g_\phi(y,\theta)) \right] ,
\end{equation}

\begin{wrapfigure}[14]{r}{0.45\textwidth}
    \centering
    \vspace{-15pt}
    \includegraphics[width=\linewidth]{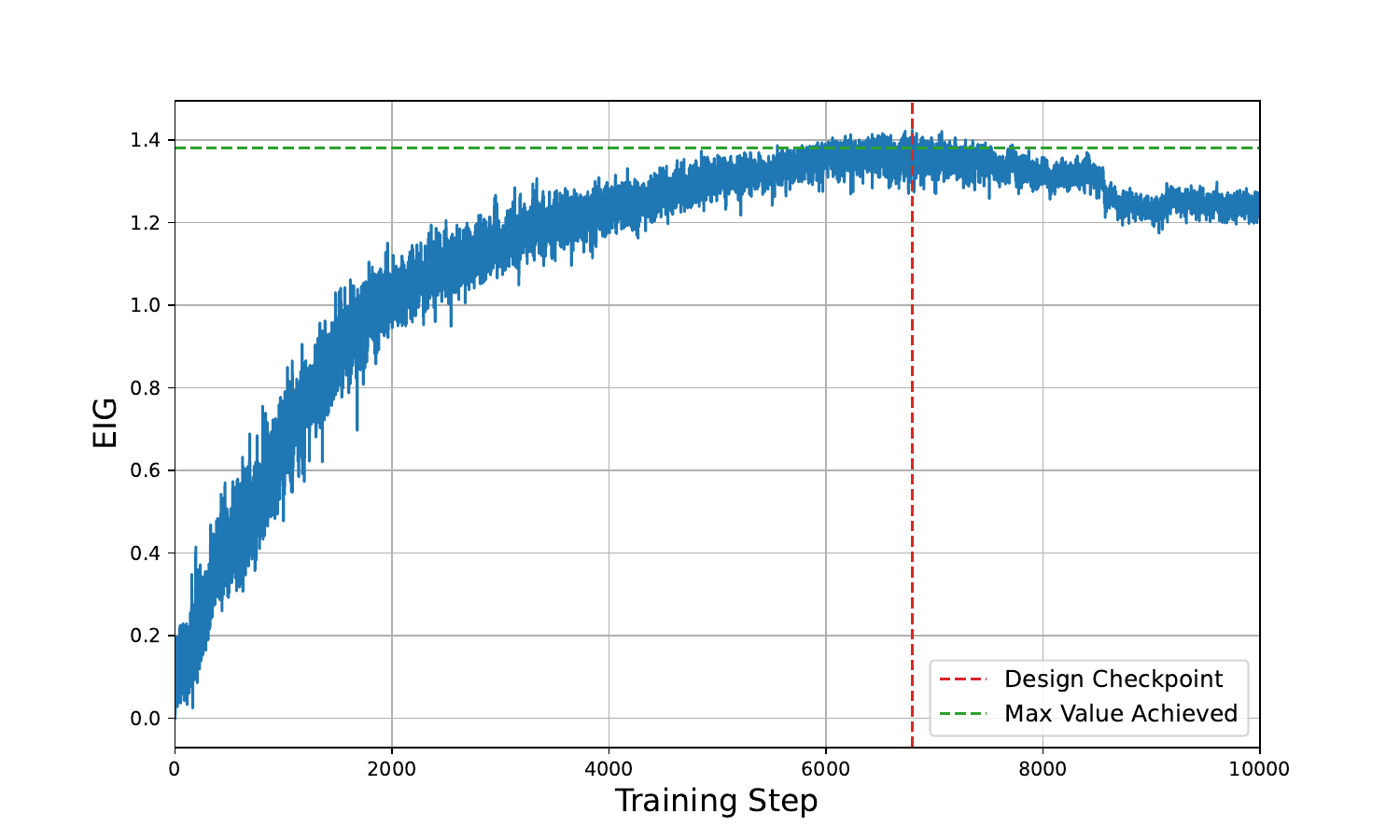}
    \caption{Design checkpointing stores the design $\xi^*$ with the highest EIG during training.}
    \label{fig:design_chkpt}
\end{wrapfigure}

where $g_\phi$ is a classifier that returns the probability that $y$ belongs to $\theta$. This function has lower bias than the InfoNCE bound but higher variance \cite{poole2019variational, song2019understanding}. In practice, this can be calculated by drawing $N$ samples from the joint distribution and shuffling to return $N (N - 1)$ marginal samples. \cite{miller_contrastive_2022} noted the connection between \cref{eq:nre} and the NWJ bound, and gave a tighter bound on the NWJ-based bound using their SBI-based CRE method. Their results hint at using a bound on the MI to optimize a likelihood-to-evidence ratio, and saw increasing EIG (lower bound of MI) with increasing number of contrastive samples. This is evident from \cref{eq:NWJ_objective}. While the NWJ bound may have less bias than the InfoNCE bound, it has higher variance that grows exponentially with the value of the true MI \cite{song2019understanding}. In the SBI setting, choosing when to use the InfoNCE or the NWJ bound will depend on what type of density estimator is required for the scientific task and ease of drawing samples from the joint distribution (simulator efficiency).

\subsection{Design Checkpoints} 
\label{sec:design_dist_appendix}

\begin{wrapfigure}[23]{r}{0.35\textwidth}
    \centering
    \vspace{-40pt}
    \includegraphics[width=\linewidth]{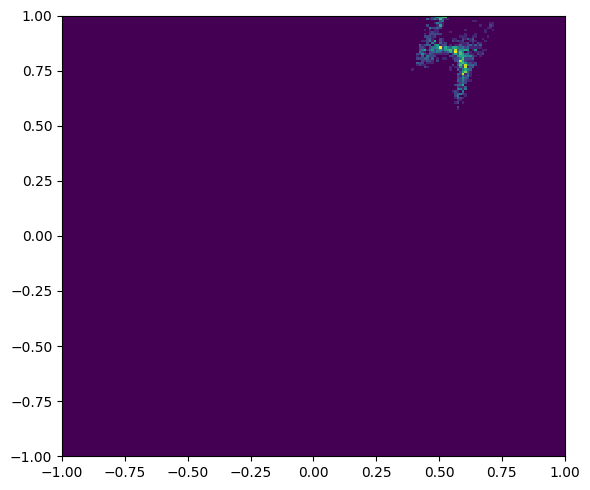}
    \vspace{0.1cm}  
    \includegraphics[width=\linewidth]{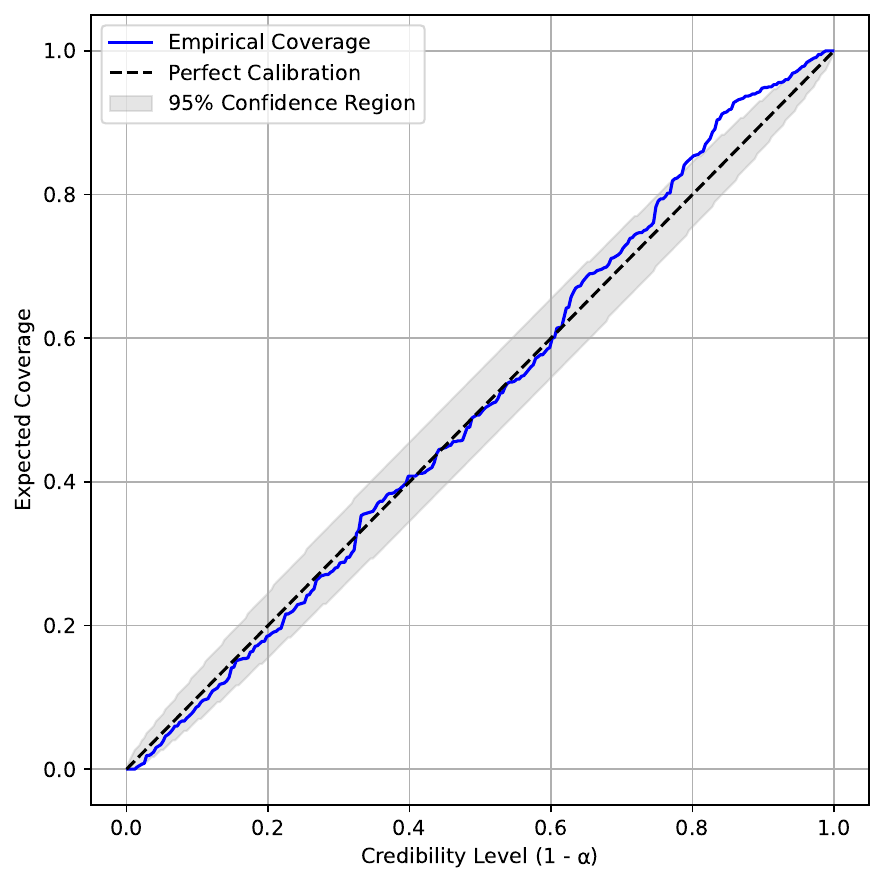}
    \caption{(Top) Posterior of the two moons experiments with mode collapse. (Bottom) Simulation-Based Calibration (SBC) of the posterior distribution for two moons.}
    \label{fig:2moons_stacked}
    \vspace{-50pt}
\end{wrapfigure}
\vspace{-10pt}

Since designs and model parameters calculate the ``reward'' in the form of the EIG, it is possible that the EIG may fall into a local optima by the end of training. This may be because of decaying learning rates of the design gradients or decreasing area searched by the design distribution. We found checkpointing mitigated this issue.

\section{Experiment Methodological Details}
\label{sec:hyperparame}

For all experiments, we use a neural spline flow (NSF) normalizing flow. We specify the different parameterizations in Table \ref{tab:hyperparams}. We used the \texttt{ReduceLROnPlateau} function to reduce the learning rate for the SIR and BMP experiments, whereas we used a constant learning rate for the linear experiment. We use the same learning rate for all flow parameters, $\phi$. Notably, for the iterated experimental design, we keep the previous round's normalizing flow parameters, $\phi_{t-1}$ to use to return the subsequent round's posterior. We set the Adam optimizer $\beta_2$ to be 0.95 to mitigate large jumps in $\xi$ that might destabilize training. All code is available online.


\subsection*{SIR Experiment Details}

We follow the implementation of \cite{ivanova2021implicit} for the SIR model. We solve an SDE describing the process using the Euler-Maruyama method and discretize the domain $\Delta \tau = 10^{-2}$. The total population is fixed at $N=500$. For the model parameters $\beta$ and $\gamma$, we use log-normal priors such that $p(\beta) = \text{Lognorm}(0.50, 0.50^2)$ and $p(\gamma) = \text{Lognorm}(0.10, 0.50^2)$. Since solving the SDE is time-consuming, we pre-simulate data on a time grid in each round and access the relevant data regions during training.

\subsection*{BMP Experiment Details}

The BMP signaling pathway can be described by mass action kinetics of proteins binding to one another and conservation laws to describe the process of a downstream genetic expression signal reaching a steady-state based on receptors available and ligands in a cell's environment. The one-step model of BMP signaling was originally proposed by \cite{su2022ligand}. While the model is described by an ODE in \cite{antebi2017combinatorial}, its steady-state signal is solved by convex optimization \cite{dirks2007thermodynamic} as a closed-box solver.

\section{Additional Two Moons Results}
\label{sec:two_moons_reloaded}

\subsection{MI optimization and mode collapse}

We analyze the mode collapse of the likelihood-based two moons posterior prediction related to the MI optimization. We also show a Simulation-Based Calibration (SBC) \cite{talts2020validating, hermanscrisis} curve for the two moons plot. For SBC we average over the parameters and compare the average posterior parameter values to the average prior values. 


\begin{figure*}[tb]
  \centering
  \includegraphics[width=\textwidth]{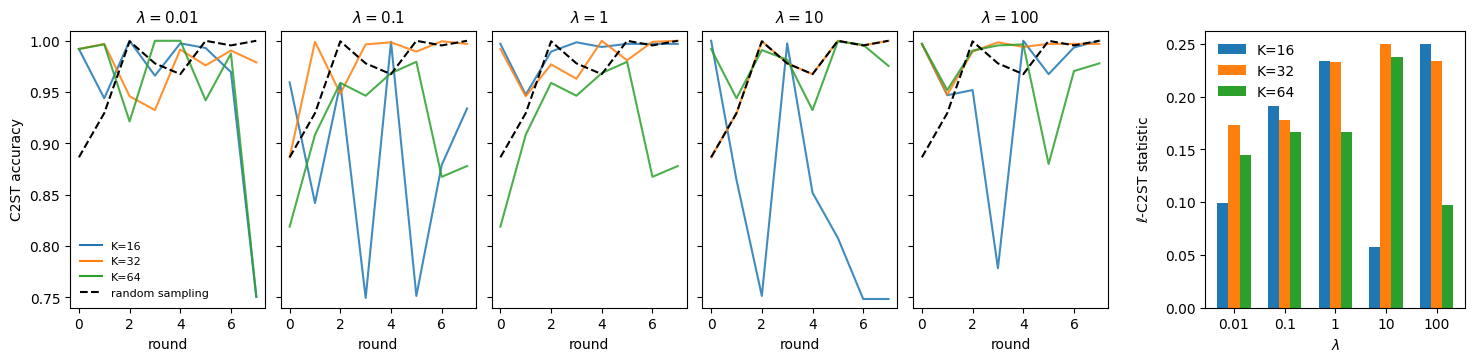}
  \vspace{-13pt}
  \caption{\textit{Left:} C2ST (lower is better) trajectories over sequential rounds for a sweep of regularization strengths $\lambda \in \{10^{-2},10^{-1},10^{0},10^{1},10^{2}\}$ (one panel per $\lambda$); curves correspond to the size of the $\theta^*$ set $K \in \{16,32,64\}$. \textit{Right:} Final $\ell$-C2ST statistic for the same sweep, grouped by $K$. Overall, larger $\lambda$ tends to increase the final $\ell$-C2ST statistic (worse calibration), while varying $K$ yields non-monotone changes, consistent with $K$ primarily affecting the variance of the epistemic approximation rather than the dominant calibration--regularization tradeoff.}
  \label{fig:two_moons_c2st_lc2st}
  \vspace{-12pt}
\end{figure*}

\subsection{active learning of simulations in sbi additional results}
\label{sec:two_moons_active_appendix}

Figure~\ref{fig:two_moons_c2st_lc2st} provides a closer look at calibration diagnostics for the Two Moons task under a sweep of the objective regularization strength $\lambda$ and the top-$K$ contrastive samples collected in $\theta^*$ used to approximate model uncertainty. Across the sweep, $\lambda$ has the most pronounced effect on posterior calibration: increasing $\lambda$ generally increases the final $\ell$-C2ST statistic, indicating worse local calibration. This suggests that, in this setting, regularization of the MI objective primarily trades off information-seeking behavior against fidelity of the learned likelihood and calibrated posterior geometry.

\begin{figure*}[hb]
  \centering
  \includegraphics[width=0.65\textwidth]{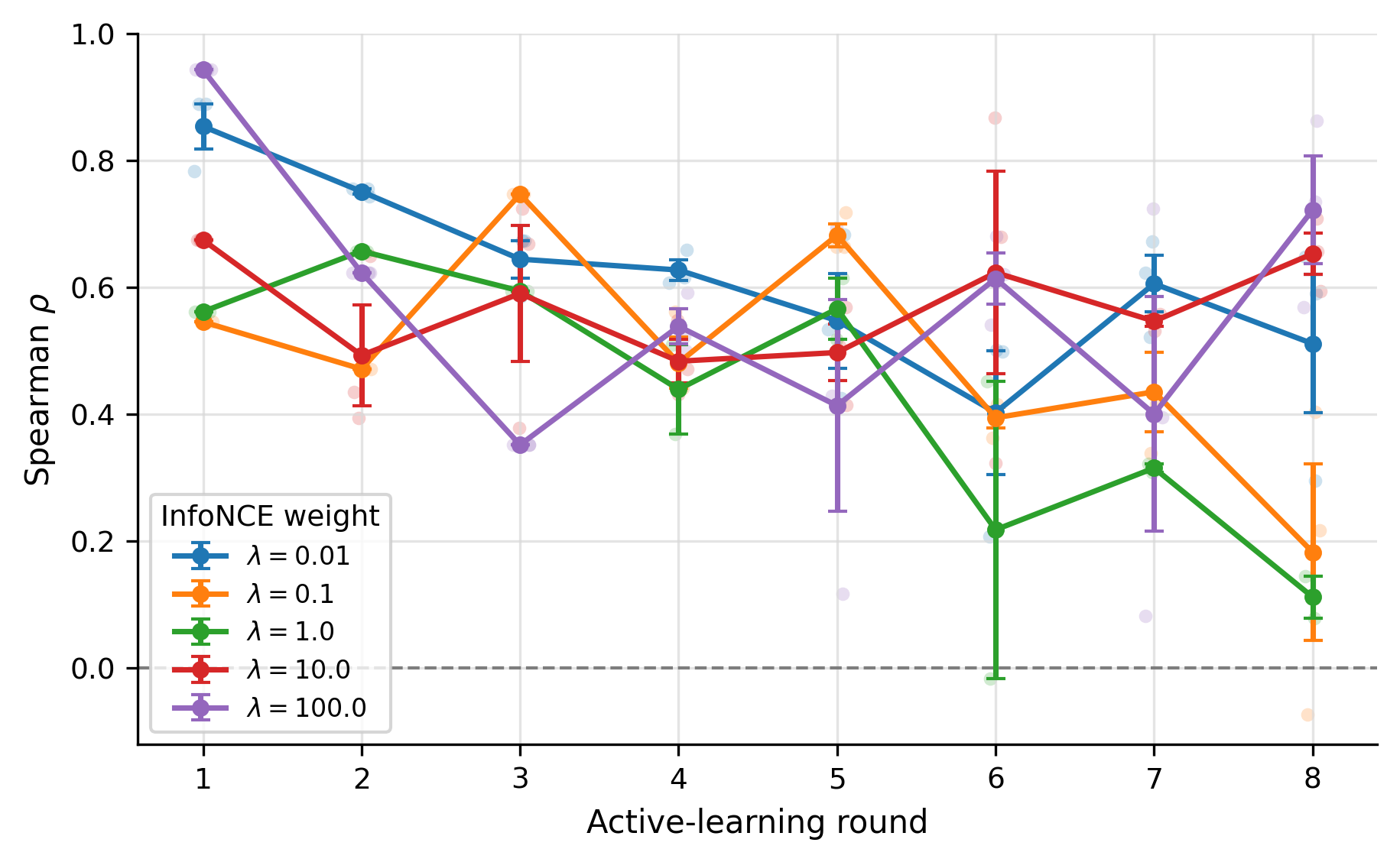}
  \vspace{-10pt}
  \caption{MC-dropout diagnostic for EPIG-based active learning on Two Moons. For each InfoNCE regularization value $\lambda$, we report the Spearman rank correlation between MC-dropout disagreement and held-out deterministic NLL over active-learning rounds; faint points show individual diagnostic runs and error bars denote standard error across the available top-$K$ settings. Positive correlations indicate that the dropout-based acquisition signal tends to rank simulator queries with higher held-out likelihood error as more uncertain.}
  \label{fig:two_moons_dropout_nll_diagnostic}
\end{figure*}

In contrast, the effect of $K$ is less systematic. While larger $K$ can improve calibration in some regimes, it does not yield a monotone improvement across all $\lambda$ values, consistent with $K$ acting mainly as a variance-control knob for the epistemic approximation rather than changing the dominant behavior induced by $\lambda$. Importantly, increasing $K$ is computationally inexpensive relative to model training, since it only increases the number of target samples to keep track of when used in the acquisition/diagnostic computations. Consequently, when compute allows, one can typically increase $K$ without materially altering the qualitative calibration--regularization trends governed by $\lambda$.

To diagnose whether MC-dropout provides a useful acquisition signal, we compare dropout disagreement against held-out likelihood error across active-learning rounds. \Cref{fig:two_moons_dropout_nll_diagnostic} shows that the dropout score is positively correlated with held-out deterministic NLL across the sweep, supporting its use as a practical uncertainty heuristic rather than a principled posterior over flow parameters.

\textbf{Mutual information optimization does not avoid mode collapse} \hspace{0.5em} Mode collapse in the two moons problem (\cref{fig:2moons_stacked}) is a common issue when using a likelihood-based flow model. While we initially considered analyzing this through the lens of mutual information and its interpretation as the expected log ratio of the posterior to prior:
$$
\mi(\theta ; y) = \E_{p(\theta, y)} \left[ \log \frac{p(y|\theta)}{p(y)} \right] = \E_{p(\theta, y)} \left[ \log \frac{p(\theta|y)}{p(\theta)} \right],
$$
our analysis in \cref{sec:MI_likelihood_proof} revealed that the root cause of mode collapse likely stems from deficiencies in the marginal likelihood approximation. There are two primary sources of error in the marginal likelihood estimation: the use of an approximate likelihood $p_\phi(y|\theta)$ and the reliance on finite samples to estimate the expectation. These approximations can introduce biases that may contribute to the observed mode collapse. Specifically, the InfoNCE bound used to maximize a lower bound of MI may be biased towards modes that better represent the observed data but potentially underrepresent the full range of parameter spaces that could explain the data.
\begin{table}[t]
\centering
\setlength{\tabcolsep}{4pt}
\renewcommand{\arraystretch}{0.9}
\caption{Training hyperparameters.}
\begin{tabular}{lcccc}
    \toprule
                                    & Linear & SIR & BMP   \\
        \toprule    
        Batch Size                      & 10   &   256  &  128  \\
        Number of Contrastive Samples     & 50   &  255 &  127  \\
        Number of Gradient Steps          &  10000  &  10000  &  500  \\
        $\phi$ \& $\xi$ Learning Rate         & $1\times10^{-3}$  &  $1\times10^{-3}$  &  $1\times10^{-3}$  \\
        Annealing Rate               & NA  &  0.8  &  0.8  \\
        Final Learning Rate               & NA  &  $1\times10^{-4}$  &  $1\times10^{-4}$  \\
        Gradient Clipping Threshold   &  NA  &  5 &  5  \\
        Hidden Layer Size          &  128  &  64 &  64  \\
        Number of Hidden Layers        &  4  &  2 &  2  \\
        Number of Flow layers (bijectors)   &  5  &  5 &  4  \\
        Number of bins for NSF        &  4  &  4 &  4  \\
\end{tabular}
\label{tab:hyperparams}
\end{table}

\begin{figure*}[h]
  \centering
  \includegraphics[width=\textwidth]{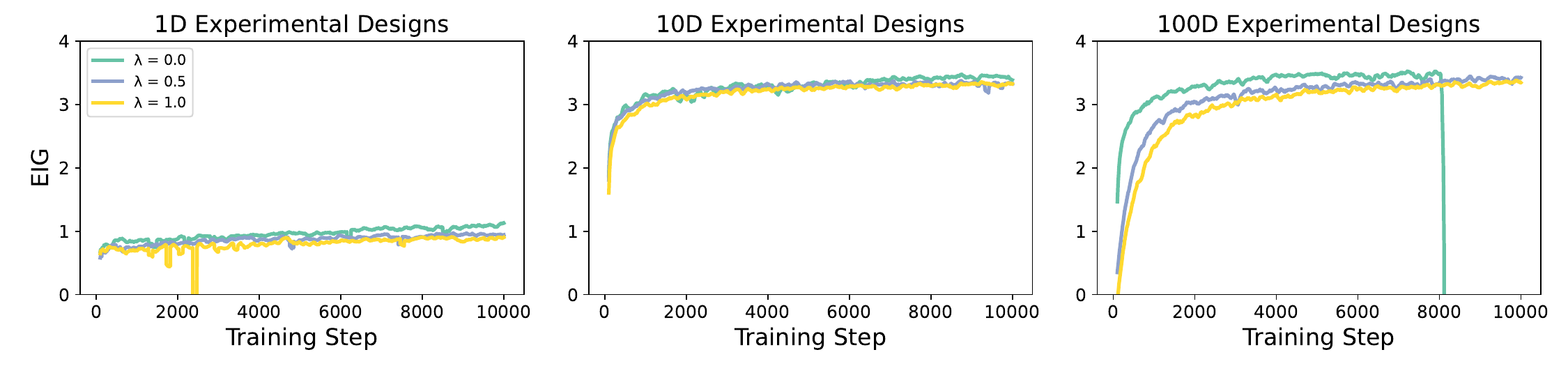}
  \vspace{-13pt}
  \caption{Comparison of the EIG across design dimensions, type of BOED, and $\lambda$ regularization for the noisy linear model examining the moving average over 10 different random seed initializations. For the single design dimension, all SBI-BOED regularization levels are generally similar, with the non-regularized version being the closest to the optimal MI bound. In the higher-dimension design cases, SBI-BOED increases its EIG with more designs. In the 100-dimensional design case, we see the benefit of using $\lambda$ regularization to stabilize the training of a design-dependent normalizing flow in high-dimensional input space at the cost of slightly lower EIG.
  } 
  \label{fig:linear_eig}
  \vspace{-12pt}
\end{figure*}

Indeed, \citet{Foster2019} suggest simultaneously optimizing a posterior distribution at the same time as a likelihood via Likelihood-Free Adaptive Contrastive Estimation (LF-ACE). This approximates the marginal likelihood with a root sample from the prior $p(\theta_0)$, and samples from an approximate posterior $\theta_i \sim q(\theta|y)$. This is essentially using samples from a posterior as importance sample estimates to help address the mode collapse of the approximate marginal likelihood. While theoretically sound, we attempted this using a normalizing flow to approximate and sample from the posterior. We found that training both the likelihood and posterior while optimizing designs to be unstable. Future work may address this with pretraining of one or both of the density estimators to improve stability of estimation.

Recent SBI work on accurate mutual information approximation \citep{miller2023simulation, glaser2022maximum} while training amortized inference networks typically relies on using a generative model and critic in a form of importance sampling similar to LF-ACE. By addressing these issues in the marginal likelihood estimation, we may be able to develop more robust methods that avoid mode collapse in the two moons problem and related scenarios.

\section{Expanded BOED Results}
\label{sec:boed_reloaded}

\begin{figure*}[h]
  \centering
  \includegraphics[width=\textwidth]{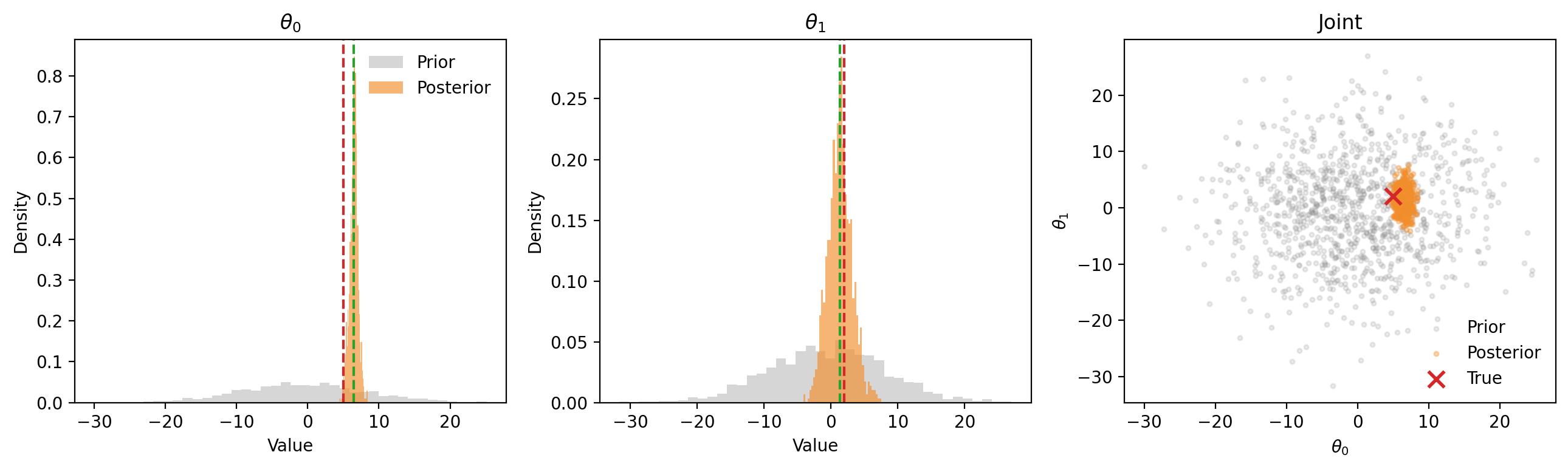}
  \vspace{-13pt}
  \caption{Posterior diagnostic for a representative 100-dimensional noisy linear design run. Marginal histograms for $\theta_0$ and $\theta_1$ and the joint posterior compare prior samples against posterior samples from the learned likelihood after optimizing the high-dimensional design; dashed lines and the red marker denote the ground-truth parameter values. The posterior contracts around the true parameters despite the 100-dimensional design vector.}
  \label{fig:linear_100d_posterior}
\end{figure*}

\textbf{Linear Model} \hspace{0.5em} We show the effect of varying design dimensions and $\lambda$ regularization on the EIG metric in \cref{fig:linear_eig}. Optimizing a likelihood while optimizing experimental designs can struggle with high-dimensional designs but this is addressed with our regularization parameter. The corresponding 100-dimensional posterior in \cref{fig:linear_100d_posterior} concentrates near the true parameters relative to the broad prior.

\textbf{Improving Posterior Estimation} \hspace{0.5em} For both the SIR and BMP model, we only use a single round of inference for each round of BOED. All SBI algorithms can be refined by sequential application of drawing posterior samples conditioned on the observed data which will then help improve the quality of the likelihood \cite{Papamakarios2018}. We forego this refinement step in our study in favor of examining how our method works with different settings of our regularization parameter. Depending on the simulator, this step can be computationally expensive but performing a calibration analysis of the likelihood can help to determine whether it is worth performing refinement steps.


\begin{figure*}[h]
  \centering
  \includegraphics[height=3.9cm, keepaspectratio]{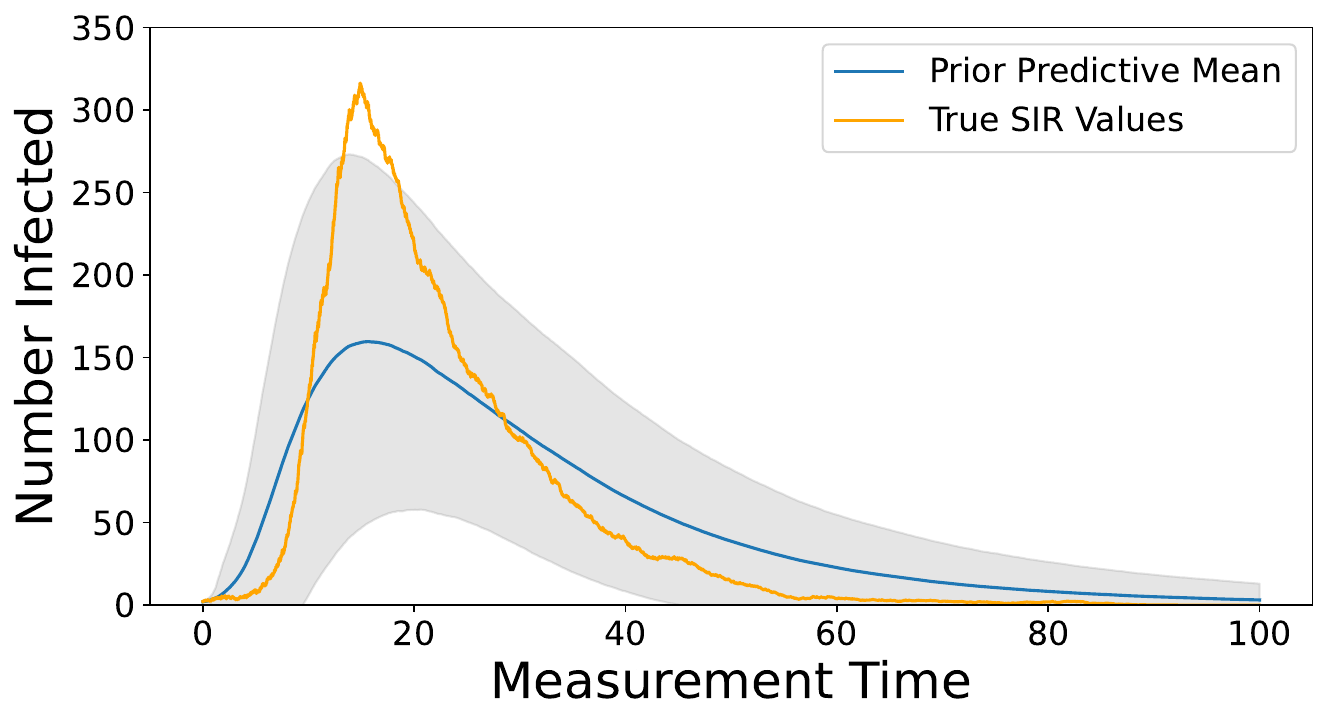}
  \hfill
  \includegraphics[height=3.9cm, keepaspectratio]{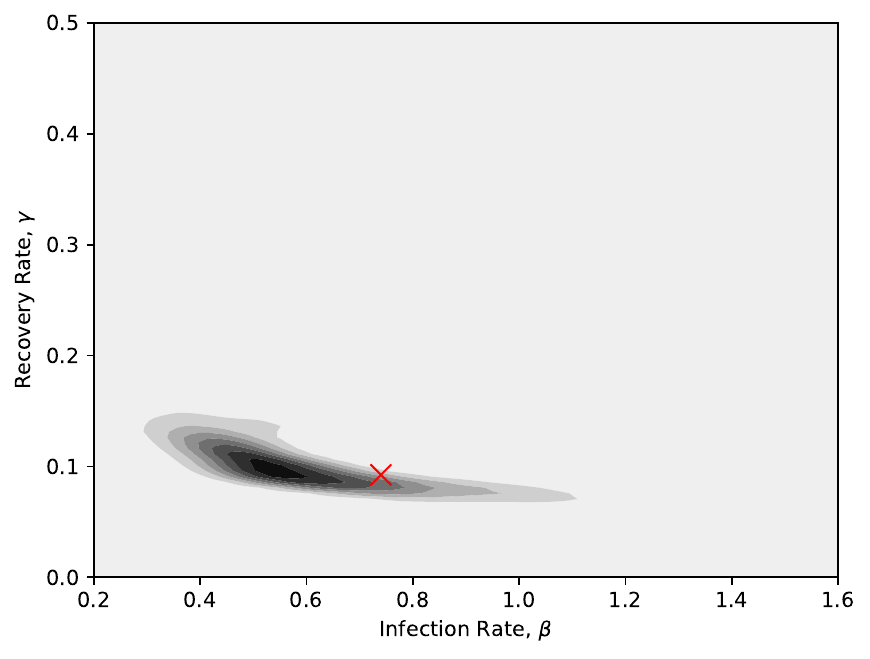}
  \hfill
  \includegraphics[height=3.9cm, keepaspectratio]{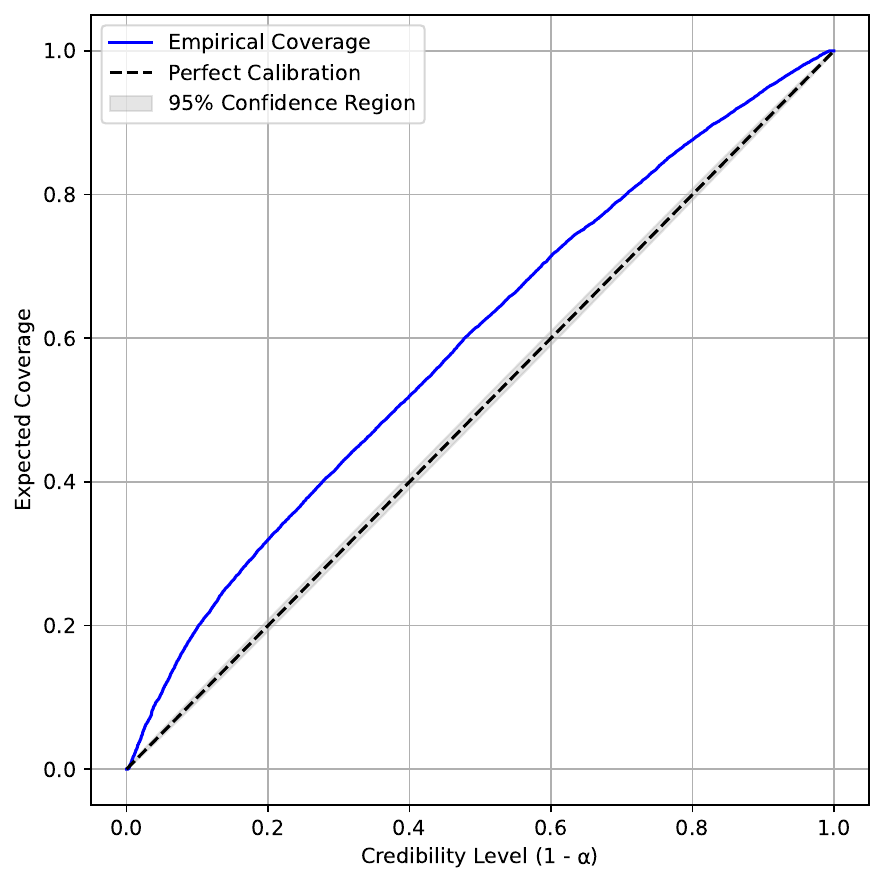}
  \caption{(\textit{Left}) Predictive SIR trajectories with the observed trajectory superimposed. (\textit{Middle}) Approximate posterior for the SIR model with true parameters marked in red. (\textit{Right}) SBC expected coverage for the SIR posterior.}
  \label{fig:sir_posterior}
\end{figure*}


\begin{figure*}[h]
  \centering
  \includegraphics[height=4.5cm, keepaspectratio]{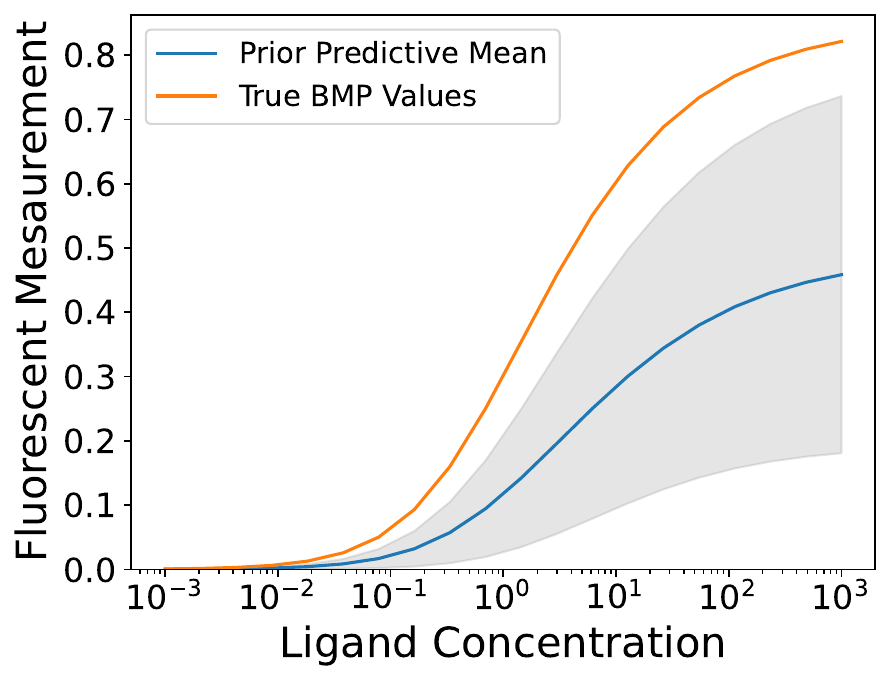}
  \hfill
  \includegraphics[height=4.5cm, keepaspectratio]{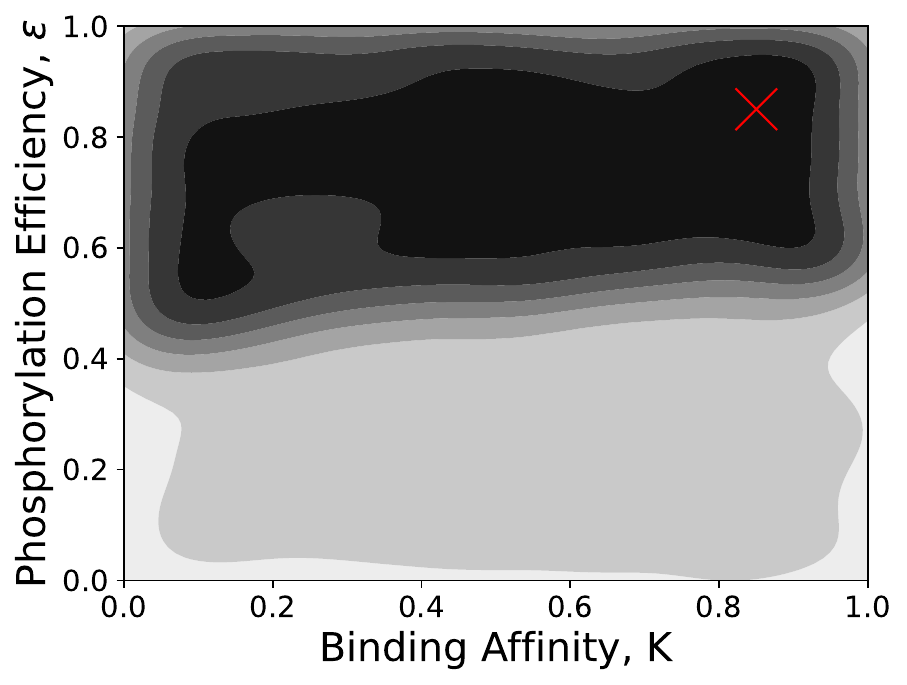}
  \hfill
  \includegraphics[height=4.5cm, keepaspectratio]{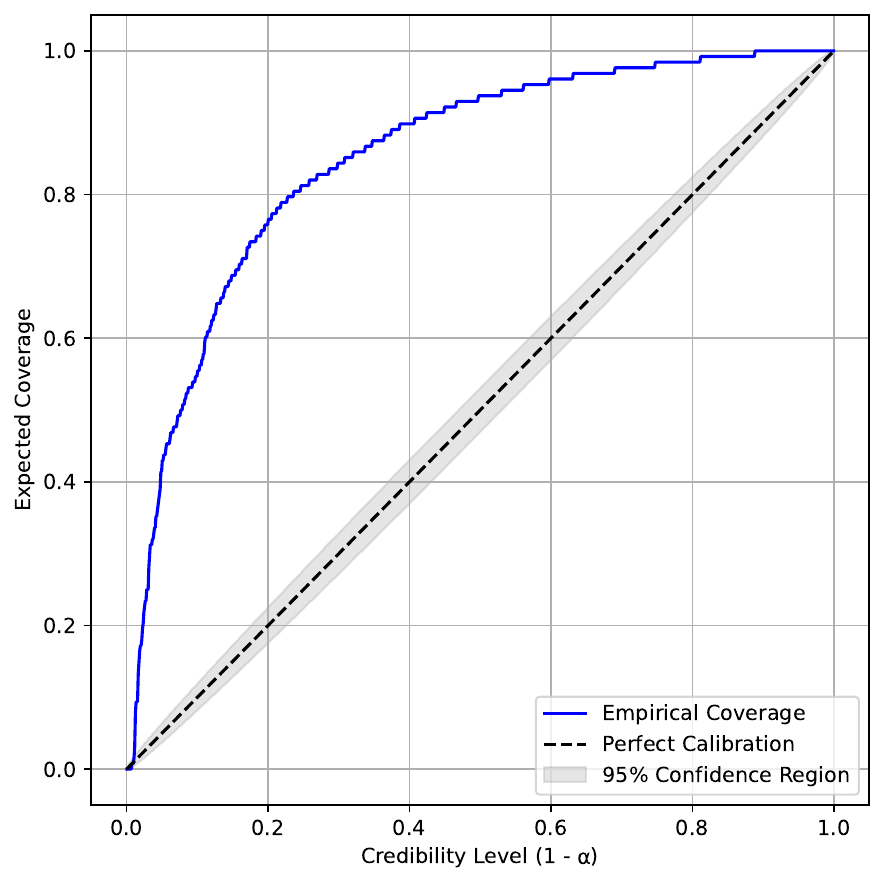}
  \caption{(\textit{Left}) Prior predictive distributions for the BMP model with the true evaluation curve superimposed. (\textit{Middle}) Approximate posterior for the BMP model with true parameters marked in red. (\textit{Right}) SBC expected coverage indicating a conservative posterior, which is supported by the posterior in question.}
  \label{fig:bmp_posterior}
\end{figure*}


\section{Applying Sequential SBI in BOED}
\label{sec:sbi_in_boed}

\begin{figure*}[htb]
  \centering
  \includegraphics[height=17cm, keepaspectratio]{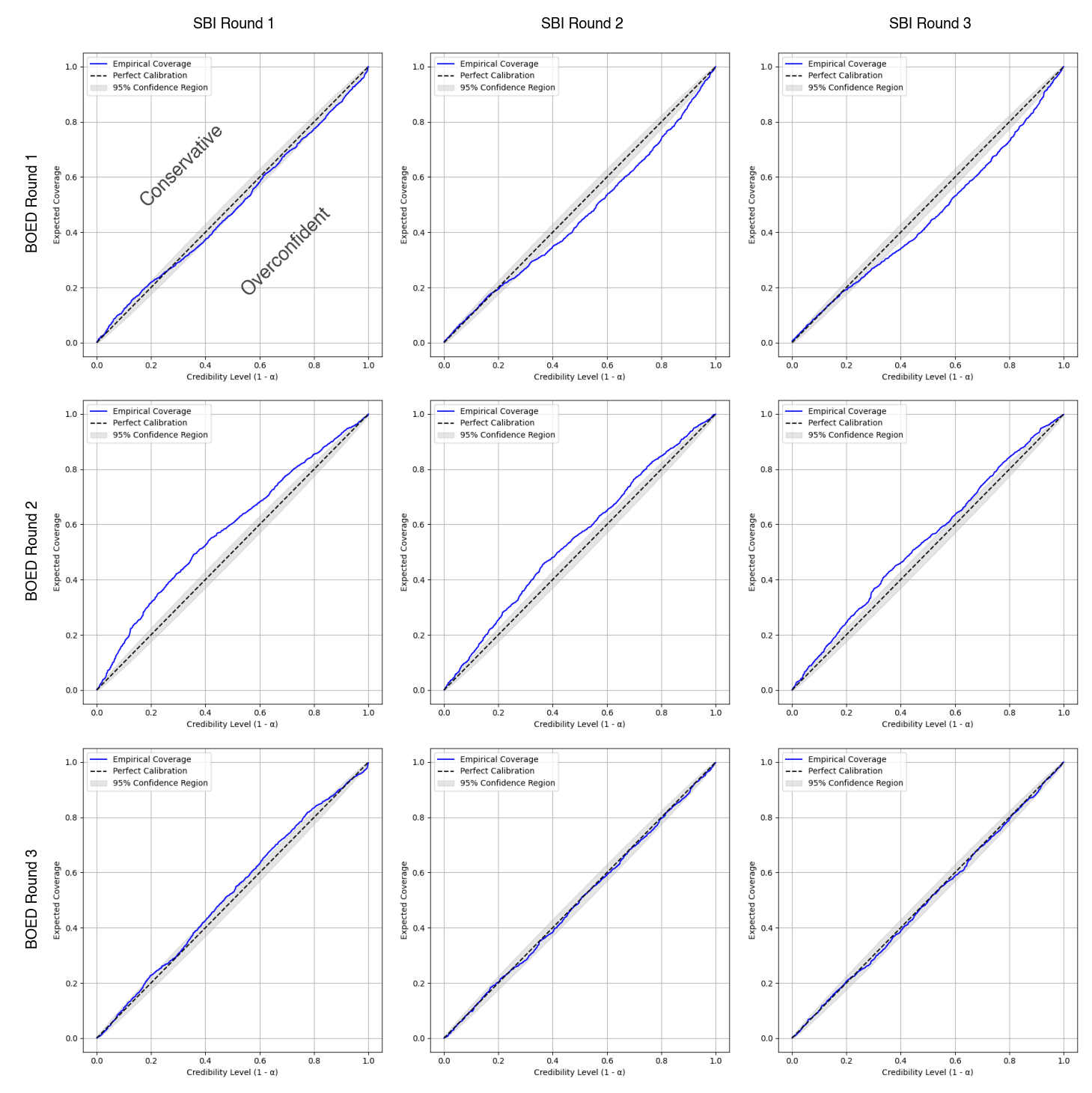}
  \caption{SBI refinement on the BMP model over 3 experimental design (BOED) rounds where the likelihood is refined after each experimental design round for three rounds of SBI. We show the SBC charts where coverage tending to the upper left indicates a conservative posterior while those to the bottom right indicate overconfident ones, and the gray area represents the 95\% confidence region. For the first round of BOED we see that refinement of the likelihood object is not necessary and performing subsequent rounds of SBI might make it overconfident. For the second BOED round, we see the initial posterior is conservative and that after two rounds of SBI becomes almost calibrated, hinting that another round of SBI might make it calibrated. The final BOED round starts slightly conservative but is calibrated after two SBI rounds. This demonstrates the modularity and flexibility in improving posterior inference that SBI methods offer in between BOED rounds.}
  \label{fig:SBI_refine}
\end{figure*}

Until now we have focused on comparing SBI-BOED to comparable BOED methods without using a key advantage of SBI methods: refinement of a likelihood, or any inference object, with observed data before designing the next experiment. Our experimental results from \cref{tab:exp_compare_expanded} show we still outperform comparable methods in accuracy and calibration, but our calibration in \cref{tab:exp_compare_expanded} indicates there is indeed room for improvement. We demonstrate the benefits of applying sequential SBI methods to refine the likelihood after every round, resulting in a calibrated posterior. This is a key benefit of SBI-BOED over alternative BOED methods that do not allow practitioners to refine their posterior distribution before designing a subsequent experiment; \cref{alg:sbi-boed-refinement} summarizes this refinement procedure. We see in \cref{fig:SBI_refine} a clear improvement of the calibration of the BMP model after the last experimental design using SBI refinement.

\clearpage
\noindent\begin{minipage}{\textwidth}
\hrule height 0.8pt
\vspace{2pt}
\refstepcounter{algorithm}\label{alg:sbi-boed-refinement}
\noindent\textbf{Algorithm \thealgorithm} SBI-BOED with likelihood refinement
\vspace{2pt}
\hrule
\vspace{2pt}
\small
\begin{algorithmic}[1]
\STATE \textbf{Require:} Simulator $p(y|\theta, \xi)$, estimator $p_\phi(y|\theta, \xi)$, prior $p(\theta)$, number of experimental designs $T$, number of BOED training steps $N$, refinement rounds $R$, simulations per refinement round $M$
\STATE \textbf{Return:} Optimal designs $\{\xi^*_i \mid i = 1, \ldots, T\}$ and refined likelihood $p_{\phi}(y|\theta, \xi)$
\STATE Initialize observed dataset $\mathcal{D}_{\mathrm{obs}} = \{ \}$
\STATE Initialize $\hat{p}_{0}(\theta) \gets p(\theta)$
\FOR{$t=1,\ldots,T$}
    \STATE Sample $\theta_{0:L} \sim \hat{p}_{t-1}(\theta)$
    \FOR{$n=1,\ldots,N$}
        \STATE Sample $\xi \sim \mathcal{N}_{\text{trunc}}(\mu_\xi \mid \sigma_n^2)$
        \STATE Simulate $y \sim p(y|\theta, \xi)$
        \STATE Estimate $\nabla \mathcal{L}_{\text{NCE}-\lambda}(\xi, \phi; L)$ via \cref{eq:lamb_lf_pce}
        \STATE Update $\mu_\xi$ and $\phi$ using $\nabla_\xi$ and $\nabla_\phi$
        \STATE Checkpoint $\xi^* = \xi$ \textbf{if} $EIG_\xi > EIG_{\xi^*}$
    \ENDFOR
    \STATE Observe $y_t$ using $\xi^*$ in an experiment
    \STATE Add $(\xi^*, y_t)$ to dataset $\mathcal{D}_{\mathrm{obs}}$
    \STATE Set $\hat{p}_{t,0}(\theta) \propto \left( \prod_{(\xi_i, y_i) \in \mathcal{D}_{\mathrm{obs}}} p_\phi(y_i \mid \theta, \xi_i) \right) p(\theta)$
    \STATE Initialize refinement dataset $\mathcal{D}_{\mathrm{sim}} = \{\}$
    \FOR{$r=1,\ldots,R$}
        \FOR{$m=1,\ldots,M$}
            \STATE Sample $\theta_m \sim \hat{p}_{t,r-1}(\theta)$
            \STATE Sample $\tilde{\xi}_m$ from $\{\xi^*_i\}_{i=1}^{t}$
            \STATE Simulate $\tilde{y}_m \sim p(y|\theta_m, \tilde{\xi}_m)$
            \STATE Add $(\theta_m, \tilde{\xi}_m, \tilde{y}_m)$ to $\mathcal{D}_{\mathrm{sim}}$
        \ENDFOR
        \STATE Refine $p_\phi(y|\theta, \xi)$ on $\mathcal{D}_{\mathrm{sim}}$
        \STATE Set $\hat{p}_{t,r}(\theta) \propto \left( \prod_{(\xi_i, y_i) \in \mathcal{D}_{\mathrm{obs}}} p_\phi(y_i \mid \theta, \xi_i) \right) p(\theta)$
    \ENDFOR
    \STATE Set $\hat{p}_{t}(\theta) \gets \hat{p}_{t,R}(\theta)$
\ENDFOR
\end{algorithmic}
\vspace{2pt}
\hrule
\end{minipage}
\clearpage

\begin{table*}[h]
\centering
\small
\setlength{\tabcolsep}{4pt}
\renewcommand{\arraystretch}{1.1}
\caption{SIR results comparing NLE and InfoNCE-$\lambda$ objectives under random and Sobol design policies. We report mean $\pm$ standard error over 3 random seeds. Higher EIG is better, while lower L-C2ST and median distance are better.}
\vspace{-6pt}
\label{tab:sir_policy_objective_sweep}
\begin{tabular*}{\textwidth}{@{\extracolsep{\fill}}l*{6}{c}}
    \toprule
    \multirow{2}{*}{\textbf{Objective}} 
    & \multicolumn{3}{c}{\textbf{Random design}} 
    & \multicolumn{3}{c}{\textbf{Sobol design}} \\
    \cmidrule(lr){2-4} \cmidrule(lr){5-7}
    & EIG $(\uparrow)$ & L-C2ST $(\downarrow)$ & Med. $(\downarrow)$ 
    & EIG $(\uparrow)$ & L-C2ST $(\downarrow)$ & Med. $(\downarrow)$ \\
    \midrule
    NLE 
    & 0.024 $\pm$ 0.022 
    & 0.0120 $\pm$ 0.0037 
    & 78.97 $\pm$ 33.23
    & 0.0033 $\pm$ 0.0013
    & \textbf{0.0015 $\pm$ 0.0005}
    & 69.04 $\pm$ 55.42 \\

    InfoNCE-$\lambda$ ($\lambda = 0.01$)
    & \textbf{0.932 $\pm$ 0.359}
    & 0.0053 $\pm$ 0.0042
    & 77.82 $\pm$ 33.09
    & 0.848 $\pm$ 0.246
    & 0.0048 $\pm$ 0.0023
    & 65.64 $\pm$ 52.06 \\

    InfoNCE-$\lambda$ ($\lambda = 0.1$)
    & 0.713 $\pm$ 0.422
    & 0.0066 $\pm$ 0.0047
    & 79.29 $\pm$ 33.42
    & \textbf{0.871 $\pm$ 0.289}
    & 0.0049 $\pm$ 0.0017
    & \textbf{59.92 $\pm$ 46.40} \\

    InfoNCE-$\lambda$ ($\lambda = 1$)
    & 0.355 $\pm$ 0.283
    & \textbf{0.0041 $\pm$ 0.0015}
    & \textbf{77.68 $\pm$ 32.59}
    & 0.304 $\pm$ 0.146
    & 0.0060 $\pm$ 0.0004
    & 68.08 $\pm$ 54.40 \\
    \bottomrule
\end{tabular*}
\vspace{-9pt}
\end{table*}

\begin{table*}[h]
\centering
\small
\setlength{\tabcolsep}{4pt}
\renewcommand{\arraystretch}{1.1}
\caption{BMP baseline sweep results comparing NLE and InfoNCE-$\lambda$ objectives under random and Sobol design policies. We report mean $\pm$ standard error over 3 random seeds. L-C2ST values smaller than $0.01$ are displayed as $0.01$ for readability.}
\vspace{-6pt}
\label{tab:bmp_policy_objective_sweep}
\begin{tabular*}{\textwidth}{@{\extracolsep{\fill}}l*{6}{c}}
    \toprule
    \multirow{2}{*}{\textbf{Objective}} 
    & \multicolumn{3}{c}{\textbf{Random design}} 
    & \multicolumn{3}{c}{\textbf{Sobol design}} \\
    \cmidrule(lr){2-4} \cmidrule(lr){5-7}
    & EIG $(\uparrow)$ & L-C2ST $(\downarrow)$ & Med. $(\downarrow)$ 
    & EIG $(\uparrow)$ & L-C2ST $(\downarrow)$ & Med. $(\downarrow)$ \\
    \midrule
    NLE 
    & 0.02 $\pm$ 0.01
    & 0.01 $\pm$ 0.01
    & 11.49 $\pm$ 5.33
    & 0.03 $\pm$ 0.02
    & 0.01 $\pm$ 0.01
    & 4.06 $\pm$ 1.15 \\

    InfoNCE-$\lambda$ ($\lambda = 0.01$)
    & \textbf{0.46 $\pm$ 0.07}
    & 0.01 $\pm$ 0.01
    & 10.86 $\pm$ 5.72
    & \textbf{0.61 $\pm$ 0.36}
    & 0.01 $\pm$ 0.01
    & 14.57 $\pm$ 6.59 \\

    InfoNCE-$\lambda$ ($\lambda = 0.1$)
    & 0.46 $\pm$ 0.10
    & 0.01 $\pm$ 0.01
    & 14.58 $\pm$ 5.14
    & 0.57 $\pm$ 0.20
    & 0.01 $\pm$ 0.01
    & 4.59 $\pm$ 2.67 \\

    InfoNCE-$\lambda$ ($\lambda = 1$)
    & 0.38 $\pm$ 0.17
    & 0.01 $\pm$ 0.01
    & 7.73 $\pm$ 5.25
    & 0.43 $\pm$ 0.20
    & 0.01 $\pm$ 0.01
    & 13.52 $\pm$ 4.24 \\
    \bottomrule
\end{tabular*}
\vspace{-9pt}
\end{table*}

\end{document}